\documentclass{article}

    \PassOptionsToPackage{numbers, compress}{natbib}
\usepackage[final]{neurips_2025}

\usepackage[utf8]{inputenc} % allow utf-8 input
\usepackage[T1]{fontenc}    % use 8-bit T1 fonts
\usepackage{hyperref}       % hyperlinks
\usepackage{url}            % simple URL typesetting
\usepackage{booktabs}       % professional-quality tables
\usepackage{amsfonts}       % blackboard math symbols
\usepackage{nicefrac}       % compact symbols for 1/2, etc.
\usepackage{microtype}      % microtypography
\usepackage[table]{xcolor}         % colors

\usepackage{multicol}
\usepackage{multirow}
\usepackage{graphicx}
\usepackage{amsmath}  % 提供 \vcenter 命令
\usepackage{booktabs} % 用于漂亮的表格线
\usepackage{wrapfig}

\usepackage{tcolorbox}
\tcbuselibrary{skins}
\usepackage{makecell}

\definecolor{darkgreen}{rgb}{0,0.5,0}
\usepackage{listings}
\usepackage{bm}

\newcommand{\name}{PolarQuant}

\title{PolarQuant: Leveraging Polar Transformation for Key Cache Quantization and Decoding Acceleration}

\author{%
  Songhao Wu$^{1}$\thanks{
    Equal contribution. 
    Work done during Songhao Wu's internship at Meituan.
  }  
  \quad Ang Lv$^{1}$\footnotemark[1] \quad Xiao Feng$^{2}$ \\
  \textbf{Yufei Zhang}$^{2}$ \quad \textbf{Xun Zhang}$^{2}$ \quad \textbf{Guojun Yin}$^{2\dagger}$ \quad \textbf{Wei Lin}$^{2}$ \quad \textbf{Rui Yan}$^{1}$\thanks{Corresponding authors: Rui Yan (\url{ruiyan@ruc.edu.cn}) and Guojun Yin (\url{yinguojun02@meituan.com}).} \\
  $^{1}$ Gaoling School of Artificial Intelligence, Renmin University of China \\
  $^{2}$ ShanghaiTech University \quad $^{3}$ Meituan \\ 
  \texttt{\{songhaowu, anglv, ruiyan\}@ruc.edu.cn} \\
  \texttt{fengxiao2023@shanghaitech.edu.cn} \\
  \texttt{\{zhangyufei08, zhangxun12, yinguojun02, linwei31\}@meituan.com}
}

\begin{document}

\maketitle

\begin{abstract}
The increasing demand for long-context generation has made the KV cache in large language models a bottleneck in memory consumption.
Quantizing the cache to lower bit widths is an effective way to reduce memory costs; however, previous methods struggle with key cache quantization due to outliers, resulting in suboptimal performance.
We propose a novel quantization approach \name, which provides a new perspective for key cache quantization and efficiently addresses the outlier dilemma.
We observe that the distribution of the key states reveals well-structured patterns under polar transformation.
Outliers generally appear in only one of the two dimensions, which are rotated together by a specific angle when rotary position embeddings are applied.
When represented as two-dimensional vectors, these dimensions exhibit well-organized patterns, with radii and angles smoothly distributed in polar space.
This alleviates the channel-wise outliers, making them well-suited for key cache quantization.
\name\ divides key vectors into groups of two-dimensional sub-vectors, encoding them as the quantized radius and the polar angle,
rather than quantizing original key vectors directly.
\name\ achieves the superior efficiency in KV cache quantization and accelerates the decoding process by turning the query-key inner product into a table lookup, all while maintaining the downstream performance of full-precision models.
Our code is available at \url{https://github.com/ericshwu/PolarQuant}.
\end{abstract}

\section{Introduction}

Large language models~(LLMs) have achieved remarkable success across a wide range of real-world applications. 
As these models evolve, the demand for enhanced long-context capabilities also increases, especially in tasks like contextual retrieval in question answering~\citep{liu-etal-2024-lost} and reasoning chains generation for complex reflection and decision-making~\citep{o12024, luo2024r1}.
However, a significant challenge in developing long-context LLMs is the rising memory cost associated with increasing context lengths, which hinders both their practical deployment and further research.

The attention mechanism~\citep{bahdanau2016neuralmachinetranslationjointly} in LLMs\footnotemark is a major source of computational overhead and memory consumption, which increase rapidly with context length.
To reduce this cost, Key-Value cache~(KV cache) is a common strategy, which stores and reuses keys and values for generation to avoid the redundant computation.
Nevertheless, as the context length increase, the memory required for KV cache can surpass that of the model weights, making it the dominant factor in overall memory usage.

\footnotetext{In this paper, we focus on decoder-only Transformer-based~\citep{transformer} LLMs using rotary position embedding (RoPE,~\citealp{rope}), which are the predominant implementation of advanced LLMs.}
Many solutions have been proposed to reduce the memory cost of KV cache.
Some studies introduce memory-efficient attention modules, such as MQA~\citep{shazeer2019mqa}, GQA~\citep{ainslie-etal-2023-gqa} and MLA~\citep{deepseekai2024deepseekv2strongeconomicalefficient}.
While promising, these module architectures require training LLMs from scratch, which limits their applicability.
Another research line focuses on the reduction of KV cache size in a compatible manner with pre-trained LLMs.
This includes techniques like KV cache eviction~\citep{cai2024pyramidkvdynamickvcache, li2024snapkv, zhang2023h2oheavyhitteroracleefficient}, which identifies and drops unimportant tokens from the cache , and KV cache quantization~\citep{zipcache, kvquant,gear,kivi,atom}.

\begin{figure*}[!t]
\begin{center}
\centerline{\includegraphics[width=\linewidth]{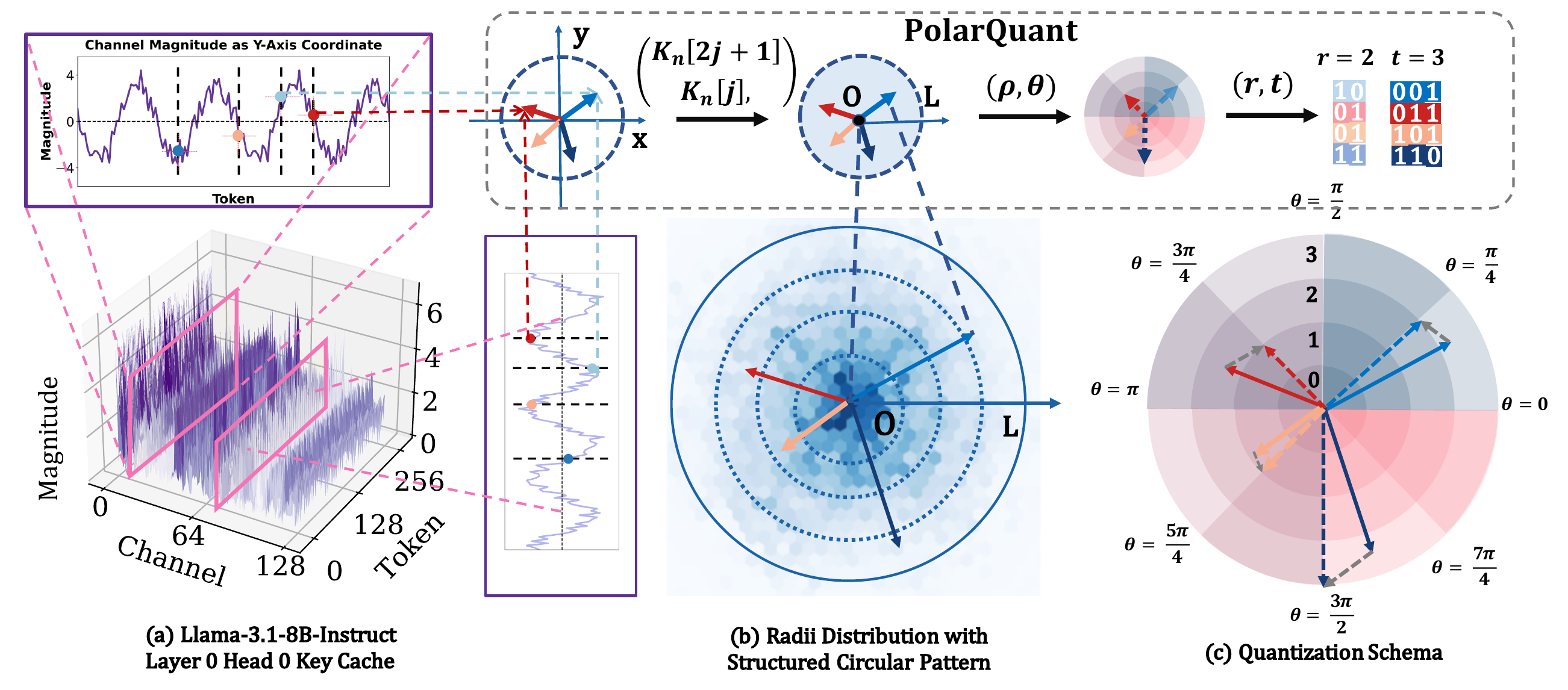}}
\caption{
% caption a typo in Figure
(a) Illustration of the activation distribution for the key cache, exemplified by Llama 3.1-8B-Instruct~(Layer 0, Head 0).
The key cache exhibits channel-wise outliers, where the magnitudes of a few channels significantly larger than others across tokens.
 We observe that these outliers generally appear in only one of the two dimensions rotated together by RoPE.
(b)
When viewing these two dimensions in a two-dimensional plane, although the individual x- or y-axis may contain outliers, they collectively form stable circular patterns, making quantization of the original outliers easier. 
Each blue dot represents a mapped two-dimensional vector, with transparency indicating frequency.
(c) \name\ using \(r=2\) bits to quantize radii and \(t=3\) bits to quantize polar angles. The colorful arrows indicate sub-vectors formed by pairs of dimensions in the keys; the quantized results are shown with colorful dashed arrows.
The quantization error is represented by the grey dashed arrow.}
\label{fig:teaser}
\end{center}
\vskip -0.1in
\end{figure*}

This paper focuses on the key cache quantization, which converts the floating-point key cache into low-bit integers to reduce memory usage.
In general, key cache quantization is more challenging than value cache due to the presence of channel-wise distributed outliers.
Prior studies~\citep{kvquant, kivi} have highlighted the widespread existence of such outliers in key cache.
As shown in Figure~\ref{fig:teaser}(a), the key states exhibit larger activations along certain channel dimensions, making token-wise quantization difficult. 
To address this issue, KIVI~\citep{kivi} proposes a channel-wise quantization strategy that groups and quantizes key elements along the channel dimensions.
% BUilding upon the channel-wise perspective
Building upon this perspective, KVQuant~\citep{kvquant} further identifies RoPE as the primary source responsible of the outliers observed in the key cache.
The rotation operations in RoPE disturb the magnitude consistency, making accurate quantization complicated.
To mitigate this, KVQuant proposes quantizing the keys before applying RoPE, which is described as pre-RoPE quantization.
Promising as it is, this approach requires on-the-fly RoPE computation, which consequently introduces potential computational overhead.
In this work, we aim to preserve the benefits of pre-RoPE quantization in reducing approximation errors while eliminating redundant computations at each generation step.
We propose a polar transformation perspective on handling outliers in the key cache, and effectively address the dilemma in 2D polar coordinates.

% As previously noted, 
KVQuant~\citep{kvquant} observes a clear and structured pattern in pre-RoPE key activations: channel-wise magnitude are highly consistent.
Recall that RoPE operates a rotation to every two-dimensional sub-vector within the key using an orthogonal $2 \times 2$ rotary matrices.\footnote{For readers unfamiliar with RoPE, please refer to Section~\ref{sec:bg}.}
Since rotation is a magnitude-preserving transformation, these 2D sub-vectors inherit the magnitude characteristics seen in the pre-RoPE case.
As shown in Figure~\ref{fig:teaser}(b), they form well-structured circular patterns when analyzed in 2D polar coordinates.
By encoding each sub-vector as its corresponding radius \(\rho\) and polar angle \(\theta\), the entire key vector can be represented as a collection of all radii and angles.
This transformation effectively mitigates outliers, as both the radii and polar angles become smoothly distributed.
Building on this, we propose a novel quantization method, \name, which significantly simplifies the quantization of the key cache.
\name\ reduces the problem of quantizing key vectors to asymmetrically quantizing $\rho$ and $\theta$ into an $r$-bit and an $t$-bit integer. 
Intuitively, \name\ defines \(2^{r+t}\) distinct regions based on \(2^{r}\) angles and \(2^{t}\) radii. 
Each sub-vector is then encoded by the index of the region it belongs to.
Figure~\ref{fig:teaser}(c) illustrates \name\ for \(r = 2\) and \(t = 3\).

\textbf{\name\ achieves superior quantization effectiveness and efficiency over previous methods.} 
On one hand, polar transformation enables smoother distributions of radii and angles, which alleviates the burden of channel-wise quantization outliers. 
The superior performance on downstream tasks further demonstrates \name's superiority in quantization error reduction.

On the other hand, \textbf{\name~offers a brand new perspective on key cache quantization, which enables a novel  decoding acceleration method.}
Unlike pre-RoPE quantization like~\cite{kvquant}, \name~eliminates the overhead of RoPE recomputation.  
% (2) Smoother distributions of radii and angles facilitate downstream performance preservation, so \name\ does not require the token grouping used in previous post-RoPE quantization~\cite{kivi}.  
% (3) \name\ also requires fewer quantization parameters, not only because it does not use grouping, but also because it leverages the non-negativity of the radii to avoid storing zero-points.
% Additionally, \textbf{\name\ enables a novel  decoding acceleration method.} 
In the attention mechanism, it replaces the standard query-key multiplication with inner products between two-dimensional query sub-vectors and a quantized polar coordinate representation of key sub-vectors, which have finite and deterministic states.
This transforms matrix multiplication to a table lookup, greatly speeding up attention computation.
% Although this approach can be applied to previous post-RoPE quantization methods, the increased number of quantization states from token grouping negates any overall efficiency gain.
% We implement Triton~\cite{triton} kernels for \name\ and our new decoding acceleration method. 
% %With $n$ = 4-bit, we achieve memory savings of \textcolor{red}{XX}\% and a \textcolor{red}{XX}\% faster，
% With $n=4$ bit, we achieve an up-to 1.27$\times$ speedup of query-key mulitplication on various open-source LLMs, while maintaining comparable downstream performance to previous competitive methods.
Our contributions are threefold: 

(1) We introduce polar transformation for key cache quantization for the first time and derive \name, a novel and efficient quantization approach; 

(2) We propose a new decoding acceleration algorithm as a natural byproduct of \name. We implement custom Triton kernels to perform fused dequantization and query-key multiplication, which achieves up to 3.18\(\times\) speedup in long-context generation;

(3) We conduct comprehensive experiments on tasks and models, which further demonstrate the superiority and robustness of our \name~across a wide range of model families and tasks.

\section{Background}
\label{sec:bg}
Consider a specific Transformer layer where the input hidden states to the attention block are denoted as \( \mathbf{X} \in \mathbb{R}^{L \times D} \), where \( L\) is the sequence length and \( D \) is the hidden state dimension. 
For any attention head, the \( d \)-dimensional query, key, and value vectors are obtained by applying three linear transformations to \( \mathbf{X} \). 
Specifically, for each head \( h \), the corresponding computations are as follows:
\[
\mathbf{\tilde{Q}} = \mathbf{X}\,\mathbf{W}_Q, \quad \mathbf{\tilde{K}} = \mathbf{X}\,\mathbf{W}_K, \quad \mathbf{V} = \mathbf{X}\,\mathbf{W}_V,
\]
where each \( \mathbf{W}_{*} \in \mathbb{R}^{D \times d} \), and the resulting variables have shapes of \( \mathbb{R}^{L \times d} \).

The query and key vectors are then applied with RoPE~\citep{rope} to incorporate positional information. 
For a query or key vector at position \( m \in [1, L] \), the corresponding rotary matrix \( \boldsymbol{R}_{m, \Phi} \in \mathbb{R}^{d \times d} \) is defined as:
\begin{align}
\boldsymbol{R}_{m, \Phi} &= \begin{bmatrix}
\boldsymbol{r}_{m, \phi_1} & \mathbf{O} & \cdots & \mathbf{O} \\
\mathbf{O} & \boldsymbol{r}_{m, \phi_2} & \cdots & \mathbf{O} \\
\mathbf{O} & \mathbf{O} & \cdots & \boldsymbol{r}_{m, \phi_{d/2}}
\end{bmatrix},
&
\boldsymbol{r}_{m,\phi_i} &= \begin{bmatrix}
\cos(m \phi_i) & -\sin(m \phi_i) \\
\sin(m \phi_i) & \cos(m \phi_i)
\end{bmatrix},
\label{eq:rope_matrix}
\end{align}
where \( \mathbf{O} \) is a zero matrix, and each \( \boldsymbol{r}_{m, \phi_i} \) is a \( 2 \times 2 \) orthogonal matrix. 
Here, \( \phi_i \) is typically defined as \( \phi_i = b^{-2i/d} \), where \( b \) is the hyperparameter for RoPE base. 

This formulation encodes the relative distance \( m - n \) between a query at position \( m > n \) and a key at position \( n \) into their inner product, as shown by:
\[
(\mathbf{\tilde{Q}}_m\,\boldsymbol{R}_{m, \Phi})~(\mathbf{\tilde{K}}_n\, \boldsymbol{R}_{n, \Phi})^{\top} = \mathbf{\tilde{Q}}_m\,\boldsymbol{R}_{m-n, \Phi}\,\mathbf{\tilde{K}}_n^{\top},
\]
the keys (after applying RoPE) and values, specifically \( \mathbf{\tilde{K}}_n\,\boldsymbol{R}_{n, \Phi} \) (which we abbreviate as \( \mathbf{K}_n \)) and \( \mathbf{V}_n \), are thus cached to avoid re-computation, known as the KV cache.
The KV cache leads to increased memory usage when processing long-context inputs.

Quantization is a simple but effective way to reduce the KV cache size. 
To clarify the mechanism, we provide a brief overview of key-value quantization below.
For the value at position \(n\), we follow the token-wise paradigm in~\citep{kivi} and quantize \( \mathbf{V}_n \in \mathbb{R}^{d} \) into $b$-bit, denoted as \( Q(\mathbf{V}_n) \).

For an arbitrary dimension $0 \leq j \textless d$, we have:
\[
Q(\mathbf{V}_{n}[j]) = \texttt{Clamp}\left( \left\lfloor \frac{\mathbf{V}_n[j] - z_n}{s_n} \right\rceil, 0, 2^b - 1 \right),
\]
where \(z_n = \min(\mathbf{V}_n[:])\) is the zero-point, \(s_n =\left( \max(\mathbf{V}_n[:]) - \min(\mathbf{V}_n[:])\right)\,/\,\left(2^b - 1\right) \) is the scaling factor.
The colon here denotes iteration over all dimensions, following Python indexing syntax. 
% use quad to force a new line
The function \( \texttt{Clamp}(x, min, max) \) restricts  \( x \) to integers within the range \( [min, max] \).

Outliers in key states make per-token quantization challenging, as we discussed earlier and illustrated in Figure~\ref{fig:teaser}(a).
To address this, previous approaches~\cite{kivi, kvquant} quantize key vectors channel-wise. 

For an arbitrary dimension \( j \), a quantized key \( Q(\mathbf{K}_n) \) at position \( n \) is given by:
\[
Q(\mathbf{K}_{n}[j]) = \texttt{Clamp}\left( \left\lfloor \frac{\mathbf{K}_n[j] - z_{[:]}{[j]}}{s_{[:]}{[j]}} \right\rceil, 0, 2^b - 1 \right),
\]
where the zero-point and scaling factor alternate as:
\[
z_{[:]}{[j]} = \min(\mathbf{K}_{[:]}[j]), \quad s_j = \frac{\max(\mathbf{K}_{[:]}[j]) - \min(\mathbf{K}_{[:]}[j])}{2^b - 1}.
\]
Here, the colon in the subscript denotes iteration over all token positions.

\section{Method}
We begin by presenting the key findings of the activation patterns in the key cache (Section~\ref{sec:motivation}). 
These insights serve as the foundation for our proposed quantization approach, \name\ (Section~\ref{sec:main_method}).

\subsection{Motivation}
\label{sec:motivation}
\begin{wrapfigure}{r}{0.48\textwidth}
    \vskip -0.3in
    \centering
    \includegraphics[width=0.96\linewidth]{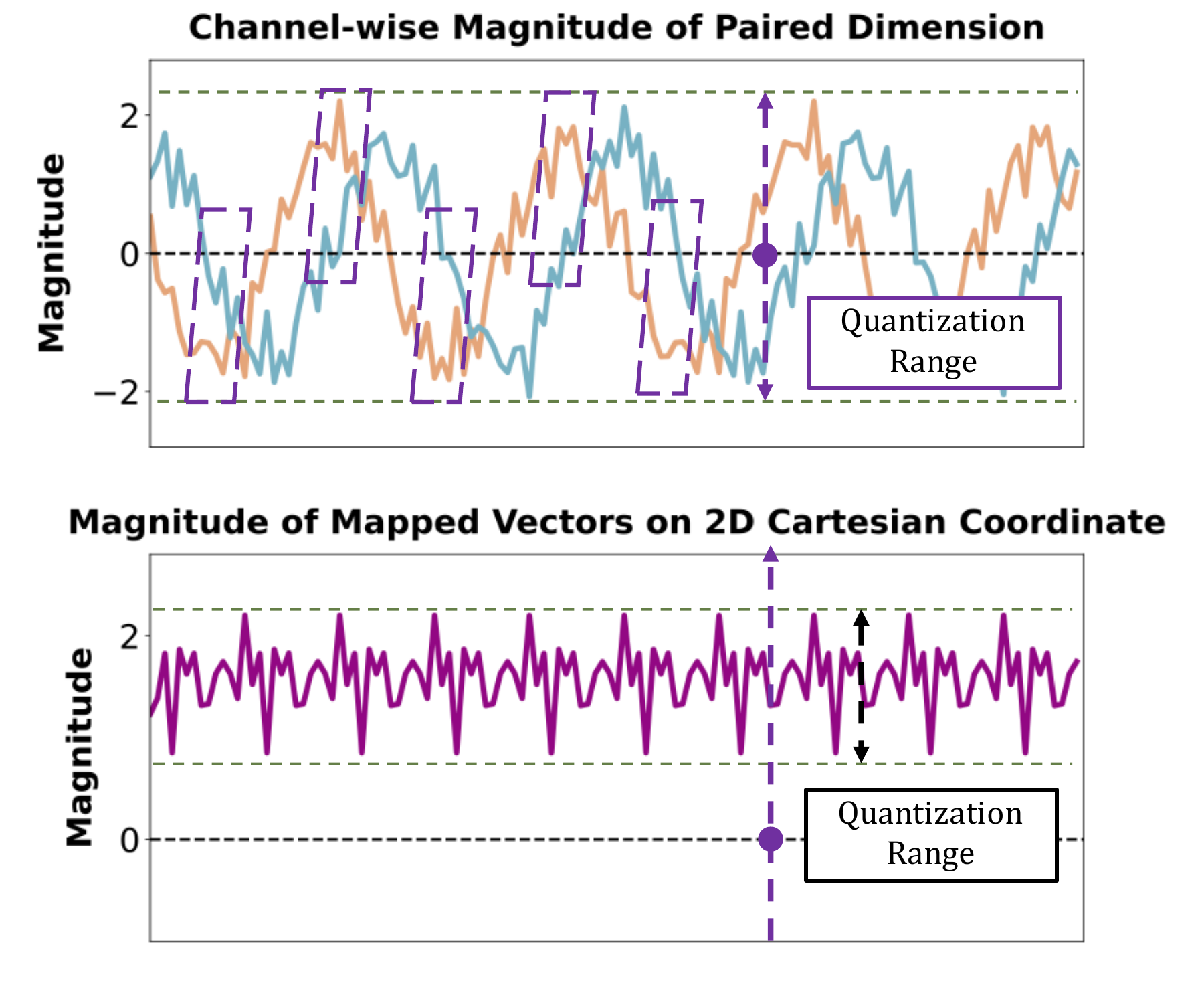}
    \caption{This figure supplements the transformation from Figure~\ref{fig:teaser}~(a) to Figure~\ref{fig:teaser}~(b), showing how \name\ better resolves the outliers.
    }
    \label{fig:motivation}
\end{wrapfigure}

As discussed earlier, outliers in key vectors pose a dilemma for key cache quantization. Our solution to this challenge arises from a key observation:

% \textrm{\textbf{Observation:}}~\textit{When mapping a paired dimension with outliers to polar coordinates, the outlier elements naturally form well-structured circular patterns, which simplify quantization.}
\textrm{\textbf{Observation:}}~\textit{When mapping the paired dimensions with outliers to polar coordinates, the resulting 2D vectors show consistent magnitudes.
The outliers present in one channel are compensated by the activations of the other, which significantly simplifies quantization.
}

Recall that in key vectors, elements in certain dimensions are jointly rotated by the same rotary sub-matrix \(\boldsymbol{r}_{n,\phi_i}\). 
Our analysis shows that the most prominent outliers 
% (highlighted in colors in Figure~\ref{fig:teaser}(a)) 
tend to occur in only one of these dimension pairs.\footnote{For efficiency, the rotary matrix is typically applied in an element-wise multiplication manner~\cite{rope}. 
To simplify implementation, dimensions \(i\) and \(i + d/2\) are often rotated together, rather than $i$ and $i+1$.
This results in non-adjacent outliers in Figure~\ref{fig:teaser}(a), but it does not affect our analysis, which is based on the matrix multiplication formulation (Eq.~\ref{eq:rope_matrix}).}

% Because items in such paired dimensions are treated as two-dimensional vectors and rotated jointly, we are motivated to analyze these outliers in a two-dimensional plane.
Figure~\ref{fig:teaser}~(b) maps the paired dimensions from Figure~\ref{fig:teaser}~(a) onto a 2D Cartesian coordinate system, where the x-axis represents the first dimension and the y-axis represents the second. 

Despite large variations in individual x and y values (which would indicate outliers in isolation), the mapped vectors form a well-structured pattern. 
In other words, when transformed into polar coordinates, the outliers are characterized by a smoothly distributed radial coordinate \(r\) and a polar angle \(\theta\). 
This structure significantly alleviates the quantization challenges faced by key caches.

Figure~\ref{fig:motivation} provides a supplementary illustration of how this polar transformation improves quantization.
The top figure illustrates a pair of dimensions exhibiting outliers, each with a large individual value range.
Quantizing this large range inevitably results in a loss of precision.
In the bottom figure, by combining these dimensions into a 2D vector, the norm (i.e., the polar radius) shows a significantly narrowed value range, facilitating quantization.

\subsection{\name: Polar-coordinate-based quantization of post-RoPE key cache}
\label{sec:main_method}

Building on these insights, we introduce a novel key cache quantization approach based on polar transformation.
Since the benefits of adopting a polar-coordinate have been outlined in the previous subsection, here we focus on the implementation details.  

For an arbitrary 2-dimensional subvector \(\Big( \mathbf{K}_{n}[2j], \mathbf{K}_{n}[2j+1] \Big) \) in the key cache at position \( n \), where \(0 \leq j < d/2 \), we interpret \(\Big( \mathbf{K}_{n}[2j], \mathbf{K}_{n}[2j+1] \Big) \) as Cartesian coordinates in the \( xy \)-plane. 
This 2D vector is then converted to polar coordinates, where the radius \( \rho_n[j] \) is given by:
\[\rho_{n}[j] = \sqrt{{\mathbf{K}_{n}[2j]}^2 + {\mathbf{K}_{n}[2j+1]}^2}, \] 
and the polar angle is:
\[ \theta_{n}[j] = \texttt{atan2}\left({\mathbf{K}_{n}}[2j+1], {\mathbf{K}_{n}[2j]}\right) + \pi
,~~\theta_{n}[j] \in \left(0, 2\pi\right),\]
where $\texttt{atan2}\left(y, x\right)$ returns the angle between the positive x-axis and the point $(x, y)$.
% , with a range of $(-\pi, \pi)$.

We perform asymmetric group-wise quantization on $\rho_n[j]$ and $\theta_n[j]$, using a group size of $g$, with $r$-bit precision for $\rho_n[j]$ and $t$-bit precision for $\theta_n[j]$:
\[
Q(\rho_{n}[j]) = \texttt{Clamp}\left( \left\lfloor \frac{\rho_n[j] - z_{\rho[:]}[j]}{s_{\rho[:]}{[j]}} \right\rfloor, 0, 2^r - 1 \right),
\]
\[
Q(\theta_{n}[j]) = \texttt{Clamp}\left( \left\lfloor \frac{\theta_n[j] - z_{\rho[:]}[j]}{s_{\theta[:]}{[j]}} \right\rfloor, 0, 2^t - 1 \right),
\]
where \(z_{\rho[:]}{[j]}\), \(z_{\theta[:]}{[j]}\) are the zero-points and \(s_{\rho[:]}{[j]}\), \(s_{\theta[:]}{[j]}\) are the scaling factors:
\[
s_{\rho[:]}{[j]} = \frac{\max(\rho_{[:]}[j]) - \min(\rho_{[:]}[j])}{2^r}, \quad
z_{\rho[:]}{[j]} = \frac{\max(\rho_{[:]}[j]) - \min(\rho_{[:]}[j])}{2^r},
\]
\[
s_{\theta[:]}{[j]} = \frac{\max(\theta_{[:]}[j]) - \min(\theta_{[:]}[j])}{2^t}, \quad
z_{\theta[:]}{[j]} = \frac{\max(\theta_{[:]}[j]) - \min(\theta_{[:]}[j])}{2^t}.
\]

Intuitively, \name~partitions the two-dimensional plane into \( 2^{r+t} \) regions, spanned by \( 2^{r} \) radii and \( 2^{t} \) polar angles. 
Each 2D sub-vector of the key vector is then represented by the center of the region in which it resides.
Figure~\ref{fig:teaser} provides an illustration of the quantization process.
The corresponding Cartesian coordinates in the key vector at dimensions \( 2j \) and \( 2j+1 \) are then calculated for the quantized representation \(\Big( Q(\rho_{n}[j]), Q(\theta_{n}[j]) \Big)\), which is formulated as:
\[
\begin{bmatrix}
\widetilde{\mathbf{K}}_{n}[2j],\;
\widetilde{\mathbf{K}}_{n}[2j+1]
\end{bmatrix}
=
\begin{bmatrix}
\tilde{\rho}_{n}[j] \cdot \cos \left( \tilde{\theta}_{n}[j] \right), \; \tilde{\rho}_{n}[j] \cdot \sin \left( \tilde{\theta}_{n}[j] \right)
\end{bmatrix},
\]
where \(\tilde{\rho}_{n}[j]\) and \(\tilde{\theta}_{n}[j]\) are the dequantized radius and polar angle:
\[
    \tilde{\rho}_{n}[j] = \big(Q(\rho_{n}[j]) + \frac{1}{2} \big) \cdot s_{\rho[:]}{[j]} + z_{\rho[:]}{[j]}, \quad \tilde{\theta}_{n}[j] = \big(Q(\theta_{n}[j]) + \frac{1}{2} \big) \cdot s_{\theta[:]}{[j]} + z_{\theta[:]}{[j]}.
\]

\subsection{Efficient decoding with \name}
% discussion of lookup table dequantization
In this section, we present \name's design for query-key multiplication, which incorporates the idea of post-multiplication dequantization to achieve  acceleration of the decoding process.
We begin by reviewing the conventional approach to dequantized generation.
Specifically, during the generation phase, the cached keys must be dequantized before being multiplied by the current query \( Q_m\) at position \( m \).
For each dimension \( 0 \leq j < d \), we have:
\[
    \bm{\widetilde{{\mathbf{K}}}}_n[j] = Q(\mathbf{K}_{n}[j])\,\cdot\,s_{[:]}[j] + z_{[:]}[j],
\]
where \( \bm{\widetilde{{\mathbf{K}}}}_{n} \) denotes the dequantized key,
and the inner product is then computed as \( \mathbf{Q}_{m} \cdot \bm{\widetilde{\mathbf{K}}_{[:]}} \).

The standard dequantization-then-multiplication operation introduces overhead for \name, as it demands extra computation.
We argue that this overhead is redundant.
At any dimension \( j \), the dequantized outcomes come from a finite set of size \( 2^{r+t} \).
Since the cache size far exceeds this set, it is more efficient to pre-compute and reuse the post-multiplication intermediate results using a lookup table~(LUT).
This approach of leveraging LUTs has been explored in prior studies~\citep{kvquant,pqcache}.
KVQuant adopts an LUT to dequantize the key cache and restores positional information via RoPE recomputation.
\name, in contrast, avoids RoPE recomputation by directly constructing a LUT for QK product on the fly.
This is the key insight behind how \name~accelerates decoding. 

Specifically, \name\ builds its lookup table within each channel by mapping quantized polar coordinates to Cartesian coordinates and computing the dot products with the query sub-vectors.
We implement custom Triton kernels to  perform fused dequantization and query-key multiplication for efficient GPU execution of \name.
The breakdown time analysis presented in Section~\ref{sec:efficency} further demonstrates the effectiveness of our \name~implementation.
More implementation details can be found in Appendix~\ref{sec:algorithm} and our released code.

\section{Experiment}

In this section, we evaluate the performance of \name~to highlight its \textit{effectiveness} and \textit{efficiency}. 
Our empirical results confirm that \name~can be integrated with LLMs, while maintaining near-lossless performance of generative tasks~(Section~\ref{sec:main}).
We also highlight the speedup achieved by \name~to showcase the superiority of our decoding algorithm and implementation~(Section~\ref{sec:efficency}).

\subsection{Preserving performance in quantized language and reasoning models}
\label{sec:main}

\paragraph{General setup.}
To ensure fair comparisons, we retain the value cache in full precision to avoid potential bias from its quantization in downstream tasks.
We evaluate \name~against several quantization baselines built on HuggingFace Transformers codebase~\citep{wolf-etal-2020-transformers}.
For all group-wise quantization methods, the group size is fixed at \(g=128\).
Additional details about the baseline methods and experimental setup are provided in Appendix~\ref{sec:main_setup}.
We evaluate all baselines alongside our \name~on a range of models from mainstream model families, including Llama~\citep{llama3,touvron2023llama2} and Qwen~\citep{qwen2025qwen25}.

\begin{table*}[t]
\caption{Performance comparison of quantization methods on LongBench. Cell colors reflect the degree of performance degradation compared to the backbone model. QJL results for Qwen2.5 are excluded due to incompatibility with its official kernel. Parenthetical values in the last column denote the performance drop of the quantized model relative to its backbone.}
\label{tab:longbench}
\vskip 0.1in
\resizebox{\textwidth}{!}
{
    \begin{tabular}{ccccccccccc}
    \toprule
     \multicolumn{2}{c}{ }  & \multicolumn{3}{c}{\textbf{Single Doc. QA}} &  \multicolumn{3}{c}{\textbf{Multi Doc. QA}} & \multicolumn{2}{c}{\textbf{Code Completion}} &  \multicolumn{1}{c}{} \\
     \cmidrule(lr{2pt}){3-5} \cmidrule(lr{2pt}){6-8} \cmidrule(lr{2pt}){9-10} \cmidrule(lr{2pt}){10-10} 
     \textbf{Quantization} & \textbf{Bits} & \textbf{NtrvQA} & \textbf{Qasper} & \textbf{MF-en} & \textbf{2Wiki} & \textbf{Hotpot} & \textbf{Musique} & \textbf{Lcc} & \textbf{RepoBench} & \textbf{Avg.~$\uparrow$} \\
     \midrule
    \multicolumn{2}{c}{ } & \multicolumn{9}{c}{\textbf{\textit{Qwen-2.5-1.5B-Instruct~(128K)}}} \\
     \cmidrule(lr{1pt}){1-2} \cmidrule(lr{1pt}){3-10} \cmidrule(lr{1pt}){11-11}
     \textbf{\textit{Bf16}} & \textbf{16} & 19.44 & 37.22 & 49.69 & 32.68 & 41.57 & 23.99 & 41.86 & 48.55 & \cellcolor{blue!10}\textbf{38.88} \\
     \midrule
     \textbf{\textit{Int-4}} & \textbf{4.25} & 5.11 & 8.80 & 10.60 & 11.99 & 5.90 & 32.82 & 24.65 & 22.57 & \cellcolor{blue!30}\textbf{15.30~(-23.58)} \\
     ${\textbf{\textit{ZipCache}}_{4}}$ & \textbf{4.25} & 4.35 & 8.52 & 12.07 & 14.14 & 17.04 & 7.57 & 23.68 & 21.48 & \cellcolor{blue!30}\textbf{13.61~(-25.27)} \\
     \textbf{\textit{KIVI-4}} & \textbf{4.25} & 19.89 & 36.52 & 49.83 & 32.18 & 41.51 & 22.89 & 40.69 & 48.72 & \cellcolor{blue!20}\textbf{36.53~~(-2.35)} \\
     ${\textbf{\textit{\name}}}_{44}$ & \textbf{4.25} & 19.34 & 36.48 & 50.80 & 32.80 & 43.25 & 23.25 & 38.62 & 46.76 & \cellcolor{blue!20}\textbf{36.41~~(-2.47)} \\
     \midrule
     \textbf{\textit{Int-3}} & \textbf{3.25} & 3.37 & 6.96 & 8.31 & 9.53 & 10.79 & 3.64 & 22.32 & 20.69 & \cellcolor{blue!35}\textbf{10.70~(-28.18)} \\
     ${\textbf{\textit{ZipCache}}_{3}}$ & \textbf{3.25} & 5.28 & 7.81 & 9.38 & 10.30 & 12.66 & 6.36 & 26.53 & 22.72 & \cellcolor{blue!35}\textbf{12.63~(-26.25)} \\ 
     \textbf{\textit{QJL}} & \textbf{3.13} & \multicolumn{9}{c}{\cellcolor{gray!20}\textbf{\textrm{N.A}}}  \\
     \textbf{\textit{KIVI-2}} & \textbf{3.00} & 18.39 & 36.07 & 47.94 & 32.51 & 42.09 & 23.33 & 39.50 & 45.31 & \cellcolor{blue!25}\textbf{35.64~~(-3.24)} \\
     ${\textbf{\textit{\name}}}_{33}$ & \textbf{3.25} & 19.09 & 35.60 & 49.47 & 32.16 & 43.47 & 23.02 & 35.77 & 44.16 & \cellcolor{blue!25}\textbf{35.34~~(-3.54)} \\
     \midrule
    \multicolumn{2}{c}{ } & \multicolumn{9}{c}{\textbf{\textit{Llama-2-7B-Chat~(4K)}}} \\
     \midrule
     \textbf{\textit{Bf16}} & \textbf{16} & 18.95 & 21.14 & 37.70 & 30.64 & 27.81 & 6.94 & 58.30 & 52.17 & \cellcolor{yellow!10}\textbf{31.71} \\
     \midrule
     \textbf{\textit{Int-4}} & \textbf{4.25} & 18.24 & 21.79 & 37.56 & 29.80 & 26.24 & 8.83 & 57.95 & 52.70 & \cellcolor{yellow!10}\textbf{31.64~~(-0.07)} \\
     ${\textbf{\textit{ZipCache}}_{4}}$ & \textbf{4.25} & 19.53 & 19.80 & 36.35 & 31.47 & 26.35 & 8.23 & 58.91 & 51.80 & \cellcolor{yellow!10}\textbf{31.55~~(-0.16)} \\
     % \textbf{\textit{KVQuant}} & \textbf{} & & & & & & & & & \\
     \textbf{\textit{KIVI-4}} & \textbf{4.25} & 18.38 & 21.16 & 37.19 & 31.67 & 26.90 & 7.85 & 58.32 & 51.99 & \cellcolor{yellow!10}\textbf{31.68~~(-0.03)} \\
     ${\textbf{\textit{\name}}}_{44}$ & \textbf{4.25} & 18.40 & 21.37 & 35.05 & 30.18 & 27.92 & 8.59 & 58.82 & 51.95 & \cellcolor{yellow!10}\textbf{31.54~~(-0.17)} \\
    \midrule
    \textbf{\textit{Int-3}} & \textbf{3.25} & 16.51 & 21.41 & 36.59 & 29.34 & 27.59 & 9.74 & 57.75 & 51.42 & \cellcolor{yellow!20}\textbf{31.29~~(-0.42)} \\ 
    ${\textbf{\textit{ZipCache}}_{3}}$ & \textbf{3.25} & 18.73 & 20.11 & 34.32 & 28.50 & 26.53 & 8.39 & 57.51 & 51.42 & \cellcolor{yellow!25}\textbf{30.69~~(-1.02)} \\
    \textbf{\textit{QJL}} & \textbf{3.13} & 19.13 & 20.51 & 35.93 & 30.74 & 25.60 & 5.79 & 57.79 & 50.92 & \cellcolor{yellow!25}\textbf{30.80~~(-0.91)} \\ 
    \textbf{\textit{KIVI-2}} & \textbf{3.00} & 18.79 & 20.46 & 35.51 & 27.52 & 26.36 & 8.12 & 56.82 & 50.26 & \cellcolor{yellow!25}\textbf{30.48~~(-1.23)} \\
    ${\textbf{\textit{\name}}}_{33}$ & \textbf{3.25} & 19.75 & 18.26 & 35.47 & 31.15 & 26.60 & 7.68 & 58.26 & 52.41 & \cellcolor{yellow!20}\textbf{31.20~~(-0.51)} \\
     \midrule
     \multicolumn{2}{c}{ } & \multicolumn{9}{c}{\textbf{\textit{Llama-3.1-8B-Instruct~(128K)}}} \\
     \midrule
    \textbf{\textit{Bf16}} & \textbf{16} & 31.38 & 46.65 & 56.81 & 49.46 & 57.85 & 32.63 & 62.88 & 56.43 & \cellcolor{red!10}\textbf{49.26} \\
    \midrule
    \textbf{\textit{Int-4}} & \textbf{4.25} & 31.76 & 45.49 & 56.47 & 49.52 & 57.67 & 32.82 & 63.17 & 55.46 & \cellcolor{red!10}\textbf{49.05~~(-0.21)} \\
    ${\textbf{\textit{ZipCache}}_{4}}$ & \textbf{4.25} & 32.26 & 45.97 & 56.77 & 49.50 & 58.67 & 33.03 & 63.41 & 55.97 & \cellcolor{red!10}\textbf{49.45~~(+0.19)} \\
    \textbf{\textit{KIVI-4}} & \textbf{4.25} & 31.23 & 47.15 & 57.14 & 49.14 & 58.05 & 32.67 & 63.05 & 56.45 & \cellcolor{red!10}\textbf{49.36~~(+0.10)} \\
    ${\textbf{\textit{\name}}}_{44}$ & \textbf{4.25} & 31.36 & 46.78 & 56.72 & 49.47 & 58.54 & 32.23 & 63.28 & 56.73 & \cellcolor{red!10}\textbf{49.39~~(+0.13)} \\
    \midrule
    \textbf{\textit{Int-3}} & \textbf{3.25} & 29.66 & 45.07 & 55.15 & 49.79 & 58.31 & 32.73 & 60.56 & 54.80 & \cellcolor{red!20}\textbf{48.26~~(-1.00)} \\
    ${\textbf{\textit{ZipCache}}_{3}}$ & \textbf{3.25} & 31.98 & 44.70 & 55.61 & 49.16 & 58.33 & 31.53 & 61.93 & 54.19 & \cellcolor{red!20}\textbf{48.43~~(-0.83)} \\
    \textbf{\textit{QJL}} & \textbf{3.13} & 32.41 & 44.75 & 56.18 & 48.50 & 57.34 & 32.07 & 61.66 & 55.99 & \cellcolor{red!20}\textbf{48.61~~(-0.65)} \\
    \textbf{\textit{KIVI-2}} & \textbf{3.00} & 31.90 & 45.39 & 54.96 & 49.88 & 58.08 & 32.25 & 62.03 & 54.93 & \cellcolor{red!20}\textbf{48.68~~(-0.58)} \\
    ${\textbf{\textit{\name}}}_{33}$ & \textbf{3.25} & 32.49 & 46.72 & 56.56 & 49.96 & 58.33 & 32.20 & 63.45 & 56.54 & \cellcolor{red!10}\textbf{49.53~~(+0.27)} \\ 
    % ${\textbf{\textit{\name}}}_{23}$ & \textbf{2.75} & 31.66 & 44.96 & 57.02 & 48.88 & 58.11 & 31.81 & 61.97 & 55.44 & \cellcolor{red!20}\textbf{48.73~~(-0.53)} \\ 
    \bottomrule
     \end{tabular}
}
\end{table*}

\paragraph{Quantizing language models.}
In real-world applications of language models, the key-value cache often becomes the primary memory bottleneck when processing long-context inputs.
We evaluate \name\ and baseline methods on LongBench~\citep{bai2023longbench}, a widely used benchmark for long-context evaluation.
Table~\ref{tab:longbench} presents results on Qwen and Llama.
There are two advantages of \name:

% 1. \name~achieves nearly lossless performance on LongBench under 3-bit and 4-bit quantization, highlighting its effectiveness in long-context language modeling.

% \textbullet \name\ delivers the most stable performance preservation across various model backbones and quantization levels, outperforming competitive baselines.
\begin{itemize}
    \item[1.] \name\ performs robustly across different model backbones.
    Quantization is especially challenging for Qwen models, which exhibit extreme channel-wise outliers in their key cache.\footnote{Qwen2.5 language models are configured with attention bias that can introduce outliers in specific channels.}
    When applied to these models, token-wise quantization methods --- such as per-token \textrm{Int.} and \textrm{ZipCache} --- tend to collapse.
    In contrast, KIVI-4 and \name\ constrain the accuracy drop to within 10\%.
    For Llama-3.1-8B-Instruct, \name\ even improves average performance under 3-bit quantization, whereas all baseline methods, including KIVI, result in performance degradation.

\item[2.] Averaged across all evaluated settings—covering both model families and quantization precisions—\name\ achieves the best overall performance preservation.
\end{itemize}

% Qwen models, naive Int-4 and ZipCache cause significant drops in performance, whereas KIVI-4 and \name\ limit the accuracy degradation to within 10%.
% For 3-bit quantized LLaMA models, \name\ even improves the average performance, while all baseline methods result in performance declines.

% 2. \name~steadily delivers competitive quantization performance across different backbones, demonstrating robustness to variations in architecture, scale, or context length.
% Qwen models exhibt extreme channel-wise outliers in their key cache.
% The model experiences a collapse, when token-wise quantization methods (such as per-token \textit{Int.} and \textit{ZipCache}) are applied.
% In contrast, \name~demonstrates greater stability under the same conditions.

% \begin{tcolorbox}[
%     colback=blue!2,  
%     colframe=black,      
%     arc=1.5mm,           
%     boxrule=0.75pt,      
%     left=2pt, 
%     right=2pt, 
%     top=1.5pt,   
%     bottom=1.5pt,  
%     width=\linewidth    
% ]
% \textbf{\textrm{\textit{Findings} 1:}} \name~achieves nearly lossless performance on LongBench under 3-bit and 4-bit quantization, highlighting its effectiveness in long-context language modeling. 
% \end{tcolorbox}

Beyond long-context processing, we also apply \name~on LLMs to examine their generative abilities with normal-length inputs.
To explore how quantization affects the emergent capabilities of LLMs, such as chain-of-thought reasoning~\citep{wei2023chainofthought}  and in-context learning~\citep{brown2020gpt3}, we benchmark \name's performance on the 5-shot CoT GSM8K~\citep{cobbe2021gsm8k}.
The results, presented in Table~\ref{tab:gsm8k}, show that \name~effectively supports both short- and long-context inputs, without compromising performance on reasoning or knowledge-intensive tasks.

\begin{table*}[htbp]
    \caption{Evaluations on 5-shot CoT GSM8K.}
    \label{tab:gsm8k}
    \vspace{0.1in}
    \centering
    \resizebox{0.84\linewidth}{!}{
        \begin{tabular}{ccccccc}
            \toprule
            % \multicolumn{1}{c}{} & \multicolumn{6}{c}{\textrm{\textbf{5-shot GSM8k~Acc.~(Bits)}}} \\
            % \midrule
            % \multicolumn{7}{c}{\textbf{\textrm{5-shot CoT GSM8K}}} \\
            \multirow{2}{*}{\raisebox{-0.75\normalbaselineskip}[0pt][0pt]{\shortstack{\textrm{\textbf{Llama-2}}\\ \textrm{\textbf{7B-Chat}}}}} & \textbf{\textrm{Quantization}} & \textbf{Bf16} & \textbf{Int-4} & \textbf{ZipCache-4} & \textbf{KIVI-4} & ${\textbf{\name}}_{44}$ \\
            \cmidrule{2-7}
            & \textbf{\textrm{Acc. (Bits)}} & 20.92~(16) & 19.79~(4.25) & 23.12~(4.25) & 21.61~(4.25) & 22.61~(4.25) \\
            \midrule
            \multirow{2}{*}{\raisebox{-0.75\normalbaselineskip}[0pt][0pt]{\shortstack{\textrm{\textbf{Llama-3.1}}\\ \textrm{\textbf{8B-Instruct}}}}} & \textbf{\textrm{Quantization}} & \textbf{Bf16} & \textbf{Int-4} & \textbf{ZipCache-4} & \textbf{KIVI-4} & ${\textbf{\name}}_{44}$ \\
            \cmidrule{2-7}
            & \textbf{\textrm{Acc. (Bits)}} & 78.85~(16) & 76.35~(4.25) & 78.70~(4.25) & 78.32~(4.25) & 78.77~(4.25) \\
            
            \bottomrule
        \end{tabular}
    }
\end{table*}

\paragraph{Quantizing reasoning models.}
Large reasoning models (LRMs) exhibit remarkable capability in solving complex problems by long chains of thought.
Recent studies~\citep{liu2025hurts} have shown that complex tasks—such as mathematics~\citep{lightman2023math500, cobbe2021gsm8k} and reasoning~\citep{rein2023gpqa}—are sensitive to the accumulation of quantization errors.
Given that LRMs rely on both long-context processing and the generation of lengthy outputs, quantization techniques are particularly critical for their application.
Moreover, the underlying mechanisms of LRMs differ from those of LLMs, making quantization more challenging and offering valuable insights into the effectiveness of different approaches.

We apply key quantization to the distillation-based reasoning models of DeepSeek-R1~\citep{luo2024r1}.
The accuracy scores across different quantization methods and tasks are presented in Table~\ref{tab:reasoning}.
The quantized reasoning models of \name~achieve substantial improvements over the baselines, which provides strong supplementary evidence of \name's superiority.

\begin{table*}[htbp]
    \caption{Overall performance comparison of quantized DeepSeek-R1-Distill models across various reasoning benchmarks. Results for \textit{ZipCache} on Qwen2.5 are omitted due to its performance collapse. Cell colors represent the degree of performance degradation compared to the backbone model.  Parenthetical values in the last column denote the performance drop of the quantized model relative to its backbone. Best results are highlighted in bold.}
    \label{tab:reasoning}
    \centering
    \vskip 0.1in
    \resizebox{0.72\textwidth}{!}{
        \begin{tabular}{ccccccc}
            \toprule
            \multirow{2}{*}{\raisebox{-0.5ex}{\textrm{\textbf{Quantization}}}} & \multicolumn{3}{c}{\textbf{AIME}} & \multirow{2}{*}{\raisebox{-0.5ex}{\textbf{MATH}}} & \multirow{2}{*}{\raisebox{-0.5ex}{\textbf{GPQA}}} & \raisebox{-0.5ex}{\multirow{2}{*}{\textbf{Overall $\uparrow$}}} \\
            \cmidrule{2-4}
            & \textbf{AIME24} & \textbf{AIME25} & \textbf{AVG.} & & & \\
            \midrule
            \multicolumn{1}{c}{} & \multicolumn{6}{c} {\textit{\textbf{DeepSeek-R1-Distill-Qwen-1.5B}}} \\
            \cmidrule(lr{2pt}){1-1} \cmidrule(lr{2pt}){2-6}  \cmidrule(lr{2pt}){7-7} 
             \textbf{\textit{Bf16}} & \textbf{36.67} & \textbf{23.33} & \textbf{30.00} & \textbf{85.20} & \textbf{39.90} & \cellcolor{blue!10} \textbf{51.70} \\
            \midrule
            \textbf{\textit{ZipCache-4}} & \multicolumn{6}{c}{\cellcolor{gray!10}N.A} \\
            \cmidrule(lr{2pt}){2-7}
            \textbf{\textit{KIVI-4}} & 20.00 & 23.33 & 21.67 & 80.40 & 33.84 & \cellcolor{blue!30} 45.30~(-6.40) \\
            ${\textbf{\textit{\name}}}_{44}$ & 30.00 & 20.00 & 25.00 & 80.20 & 37.88 & \cellcolor{blue!20} \textbf{47.69~(-3.31)} \\
            \midrule
            \multicolumn{1}{c}{} & \multicolumn{6}{c} {\textit{\textbf{DeepSeek-R1-Distill-Llama-8B}}} \\
             \cmidrule(lr{2pt}){1-1} \cmidrule(lr{2pt}){2-6}  \cmidrule(lr{2pt}){7-7} 
            \textbf{\textit{Bf16}} & \textbf{50.00} & \textbf{36.67} & \textbf{43.33} & \textbf{91.20} & \textbf{51.52} & \cellcolor{red!10}\textbf{62.01 }\\
            \midrule
            \textbf{\textit{ZipCache-4}} & 43.33 & 43.33 & 43.33 & 91.60 & 48.48 & \cellcolor{red!15}61.13~(-0.88) \\
            \textbf{\textit{KIVI-4}} & 43.33 & 33.33 & 38.33 & 89.80 & 51.01 & \cellcolor{red!20}59.71~(-2.30) \\
            ${\textbf{\textit{\name}}}_{44}$ & 60.00 & 36.67 & 48.33 & 89.00 & 50.00 & \cellcolor{red!10}\textbf{62.44~(+0.43)} \\
            \bottomrule
            
        \end{tabular}
    }
\end{table*}

\subsection{Superior efficiency of~\name}
\label{sec:efficency}
To evaluate the efficiency of the customized decoding algorithm, we provide a comprehensive time breakdown analysis of our \name\ implementation. 
In all experiments, we use the Llama-3.1-8B-Instruct model configuration, which includes 32 query heads of dimension 128, and 8 key/value heads (grouped-query attention~\citep{ainslie-etal-2023-gqa}).
We benchmark the latency of our tailored kernel implementation, as well as the end-to-end generation throughput.

\begin{figure}[hb]
    \centering
    \includegraphics[width=0.96\linewidth]{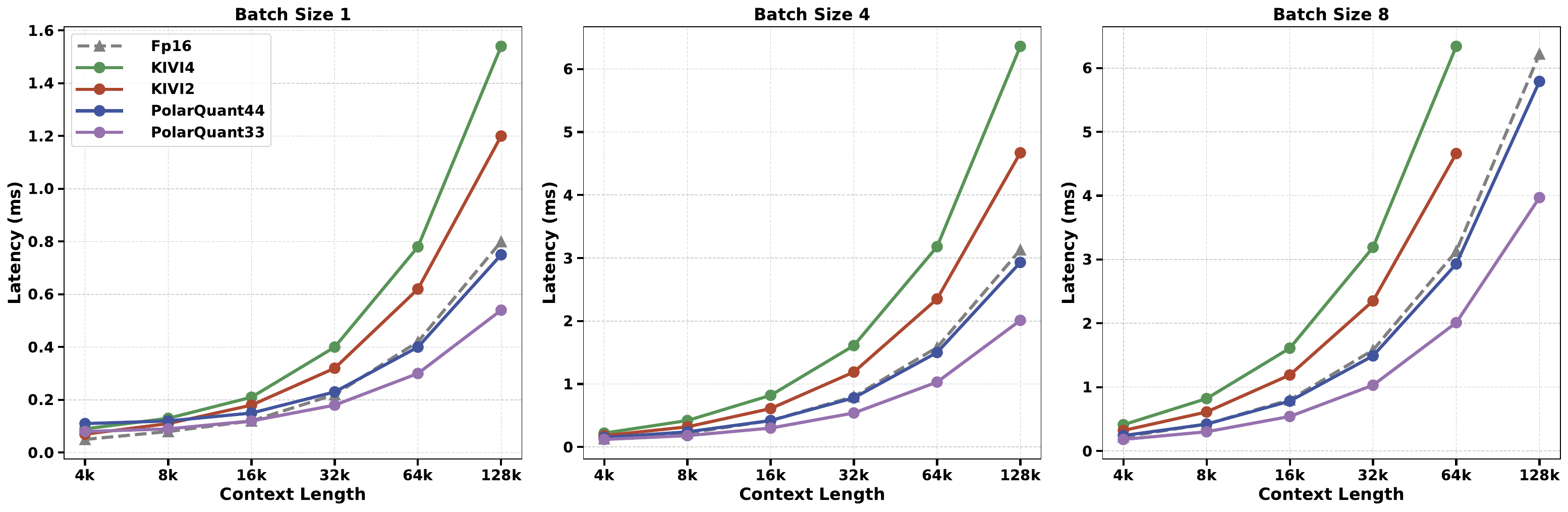}
\caption{Latency comparisons of \name~across varying batch sizes and context lengths}
    \label{fig:qk-latency}
\end{figure}

\paragraph{Latency for query-key multiplication kernel.}
We make comparisons of the query-key multiplication implementations for LLM decoding. 
Specifically, we evaluate the wall-clock latency across different sequence lengths and batch size settings. 
We benchmark the runtime by summing across 10K iterations of the calculations, and report results for: \textit{Fp16}, \textit{KIVI-4}, \textit{{KIVI-2}}, ${\textit{\name}}_{44}$ and ${\textit{\name}}_{33}$.
Figure~\ref{fig:qk-latency} illustrates the performance comparison among kernel implementations, while Table~\ref{tab:latency_thoroughtput} reports the multiplication latency with a batch size of 8.
We see that, \name~achieves up to~\textbf{2.7$\times$} speedup compared with KIVI and~\textbf{1.6$\times$} speedup compared with Fp16 Torch implementation.

\paragraph{Throughput comparison.}
We incorporate our custom kernels into the generation pipeline and measure their end-to-end performance.  
We determine the maximum supported batch size with a sequence length of 32K tokens, with Hugging Face’s implementation.
We fix the input length at 256 tokens and measure throughput across different generation lengths.
Table~\ref{tab:latency_thoroughtput} presents the results; entries marked with $\dagger$ indicate the application of 2-bit value quantization.
As is shown in Table~\ref{tab:latency_thoroughtput}, \name~achieves up to\textbf{ 3.18$\times$} throughput improvements, and demonstrates significant improvements over KIVI in both latency and throughput.  
Values in parentheses denote the speedup.

\begin{table*}[htbp]
    \centering
    \caption{Latency, throughput and memory usage comparisons.}
    \label{tab:latency_thoroughtput}
    \vskip 0.1in
    \resizebox{0.88\linewidth}{!}{ 
        \begin{tabular}{ccccc}
        \toprule
            \multirow{2}{*}{\textrm{\textbf{Operation}}} & \multicolumn{4}{c}{\textbf{\textrm{Latency~(s)}}} \\
            \cmidrule{2-5}
            & \textbf{\textrm{4K}} & \textbf{\textrm{8K}} & \textbf{\textrm{32K}} & \textbf{\textrm{128K}} \\
       \midrule
      \textbf{\textit{Fp16}} & \textbf{\textrm{0.22}} & \textbf{\textrm{0.42}} & \textbf{\textrm{1.58}} & \textbf{\textrm{6.22}}  \\
       \midrule
       \textbf{\textit{KIVI-4}} & ~0.41~($0.54\times$) & ~0.82~($0.51\times$) & ~3.19~($0.50\times$) & N.A \\ 
       ${\textbf{\textit{\name}}}_{44}$ & 0.24~($0.92\times$) & 0.42~($1.00\times$) & 1.49~($1.06\times$) & 5.79~($1.07\times$) \\ 
       \midrule
       \textbf{\textit{KIVI-2}} & 0.32~($0.69\times$) & 0.61~($0.69\times$) & 2.35~($0.67\times$) & N.A \\
       ${\textbf{\textit{\name}}}_{33}$ & 0.18~($1.22\times$) & 0.30~($1.40\times$) & 1.03~($1.53\times$) & 3.97~($1.57\times$) \\
       \bottomrule
       \addlinespace[6pt]
        \toprule
            \multirow{2}{*}{\textrm{\textbf{Configuration}}} & \multicolumn{4}{c}{\textbf{\textrm{TP.~(token/s)}}~\ / \textbf{\textrm{Mem.~(GB)}}} \\
            \cmidrule{2-5}
            & \textbf{\textrm{4K}} & \textbf{\textrm{8K}} & \textbf{\textrm{16K}} & \textbf{\textrm{32K}} \\
       \midrule
      \textbf{\textit{Fp16}} & \textbf{119.1~/~20.78} & \textbf{~84.5~/~25.21} & \textbf{~53.6~/~34.07} & \textbf{15.3~/~51.79}  \\
      \midrule
       \textbf{\textit{KIVI-4}} & 129.0~($1.09\times$)~/~19.03 & ~96.9~($1.15\times$)~/~21.88 & 65.1~($1.21\times$)~/~27.59 & 39.0~($2.54\times$)~/~38.96 \\ 
       ${\textbf{\textit{\name}}}_{44}$ & 138.8~($1.17\times$)~/~19.04 & 108.5~($1.28\times$)~/~21.90 & 75.9~($1.42\times$)~/~27.61 & 46.8~($3.05\times$)~/~39.06 \\ 
       \midrule
       \textbf{\textit{KIVI-2}} & 133.2~($1.12\times$)~/~18.89 & ~99.9~($1.18\times$)~/~21.55 & 67.7~($1.26\times$)~/~26.91 & 40.5~($2.64\times$)~/~37.62 \\
       ${\textbf{\textit{\name}}}_{33}$ & 144.8~($1.22\times$)~/~19.04 & 111.1~($1.31\times$)~/~21.90 & 78.6~($1.46\times$)~/~27.61 & 48.7~($3.18\times$)~/~39.06 \\
       \midrule
       ${\textbf{\textit{KIVI-4}}}^{\mathbf{{\dagger}}}$ & 129.0 ~($1.09\times$)~/~17.08 & 115.8~($1.37\times$)~/~17.99 & ~90.8~($1.69\times$)~/~19.92 & 30.13~($1.97\times$)~/~23.60 \\
       ${\textbf{\textit{\name}}}_{44}^{\mathbf{\dagger}}$ & 144.1~($1.21\times$)~/~17.06 & 128.0~($1.51\times$)~/~17.97 & 111.6~($2.08\times$)~/~19.81 & 46.86~($3.06\times$)~/~23.46 \\
       \bottomrule
    \end{tabular}
    }
\end{table*}

\section{Discussions}

\subsection{Ablation studies of \name}
\paragraph{Effect of group size \(g\).} \name\ applies group-wise quantization along the channel dimension.
We conduct ablation studies to investigate the impact of the group size \(g\) on model performance.

\begin{table*}[h]
    \centering
    \caption{Ablation study of group size \(g\) on LongBench.}
    \label{tab:group}
    \vskip 0.1in
    \resizebox{0.88\linewidth}{!}{
        \begin{tabular}{cccccc}
            \toprule
            & \textbf{Group Size} & 32 & 64 & 128 & 256 \\
            \midrule
            \multirow{2}{*}{\shortstack{\textbf{LongBench} \\ (Bits)}} & \textbf{KIVI-4} & 49.48~(5.00) & 49.47~(4.50) & 49.36~(4.25) & 49.52~(4.13) \\
            & ${\textbf{\name}}_{44}$ & 49.50~(5.00) & 49.33~(4.50) & 49.39~(4.25) & 49.58~(4.13) \\
            \bottomrule
        \end{tabular}
    }
\end{table*}

Specifically, we benchmark \name~and KIVI on LongBench using the Llama-3.1-8B-Instruct model.
The results, presented in Table~\ref{tab:group}, indicate that \name~performs competitively with KIVI across all tested group sizes.
It is noteworthy that the quantization parameters occupy $32/g$ bits;
therefore, smaller values of $g$ result in higher average bit widths.
To strike an optimal balance between performance and parameter overhead,
we set \(g\) as 128 throughout this paper.

\paragraph{Impact of RoPE configuration.}
We conduct experiments to test the sensitivity of \name\ to different RoPE configurations and draw the following conclusions:

(1) \name\ exhibits consistent performance across LLMs with different RoPE base frequencies~(see Table~\ref{tab:longbench}). 
The experimental results for basic values of $\{10000, 500000, 1000000\}$, highlight \name's insensitivity to the choice of base frequency.

(2) \name\ is adaptable to different RoPE variants. We employ NTK RoPE scaling~\citep{codellama} to extend the LLM's context and apply \name\ to the key cache.
Critically, mo significant performance drop is observed. 
We provide detailed experimental results and setups in Appendix~\ref{sec:rope}.

\paragraph{Bitwidth allocation for radii and polar angles.} In \name, we assign bitwidth asymmetrically between radii and angles. 
To determine which component requires higher precision, we conduct ablation studies on various bitwidth configurations, enabling a more flexible allocation strategy.
\begin{table*}[htbp]
    \centering
    \caption{Ablation study on asymmetrical bitwidth allocation in \name.}
    \label{tab:bitwidth}
    \vskip 0.05in
    \resizebox{0.88\linewidth}{!}{
        \begin{tabular}{ccccccccc}
            \toprule
            % \multicolumn{9}{c}{\textrm{\textbf{Allocation stratgies and Results}}} \\
            % \midrule
          \multirow{2}{*}{\textrm{\textbf{LongBench}}} &  \textbf{\textrm{Bits}} & \(\mathbf{(r5,t3)}\) & \(\mathbf{(r4,t4)}\) & \(\mathbf{(r3,t5)}\) &
           \textbf{\textrm{Bits}} & \(\mathbf{(r4,t2)}\) & \(\mathbf{(r3,t3)}\) & \(\mathbf{(r2,t4)}\) \\
            \cmidrule{2-9}
           & 4.25 & 49.41 & 49.39 & 49.51 & 3.25 & 47.66 & 49.53 & 49.00 \\
            \bottomrule
        \end{tabular}
    }
\end{table*}

We take Llama-3.1-8B-Instruct as backbone and benchmark \name~on LongBench, to assess the impact of different bitwidth settings.
From Table~\ref{tab:bitwidth}, we have the following observations:

\textbf{\textrm{\textit{Observation} 1:}}~Angle quantization is more sensitive to bitwidth. Allocating fewer than 3 bits to angles often results in a significant drop in performance.

\textbf{\textrm{\textit{Observation} 2:}}~Despite the asymmetry in bitwidth allocation, a more balanced distribution between radii and angles can still achieve strong performance.

\subsection{Compatibility with existing KV cache compression techniques}
\label{sec:sensitive}

In this section, we explore the integration of existing KV compression techniques with \name, which allows for further reductions in KV cache memory occupation.
We present the experimental results of Llama-3.1-8B-Instruct on LongBench. 
As stated in Section~\ref{sec:main}, we retain the value cache in full precision, as key cache is more sensitive to low-precision quantization.
Appendix~\ref{sec:apdx_sensitive} provides further analysis to support this.
We combine value quantization with \name\ to verify its compatibility. 
A standard token-wise quantization is applied to the value cache, consistent with KIVI~\citep{kivi}.
Table~\ref{tab:mixed} presents the results. The introduction of value quantization results in only marginal performance degradation, even at 2-bit precision.

\begin{table*}[htbp]
\caption{LongBench results of \name\ with value cache quantization . 
% Cell colors represent the degree of performance degradation compared to the backbone model.
}
\centering
\label{tab:mixed}
\vskip 0.1in
\resizebox{\textwidth}{!}{
    
    \begin{tabular}{ccccccccccc}
    \toprule
    \textbf{\textrm{Quantization}} & \textbf{Value Bits.} & \textbf{NtrvQA} & \textbf{Qasper} & \textbf{MF-en} & \textbf{2Wiki} & \textbf{Hotpot} & \textbf{Musique} & \textbf{Lcc} & \textbf{RepoBench} & \textbf{Avg.} \\
    \midrule
    \multirow{3}{*}{$\textbf{\textrm{PolarQuant}}_{44}$} & \cellcolor{gray!10}\textbf{16} & 31.36 & 46.78 & 56.72 & 49.47 & 58.54 & 32.23 & 63.28 & 56.73 & \textbf{49.39} \\
    \cmidrule{2-11}
     & \cellcolor{gray!10}\textbf{4} & 31.67 & 46.51 & 56.48 & 49.74 & 58.48 & 32.41 & 63.50 & 56.59 & \textbf{49.42~(+0.03)} \\
     & \cellcolor{gray!10}\textbf{2} & 31.26 & 46.69 & 57.59 & 47.88 & 58.51 & 32.98 & 63.24 & 55.95 & \textbf{49.26~(-0.13)} \\
    \bottomrule
    \end{tabular}
}
\end{table*}

We further explore the integration of \name~with token-eviction strategies~\citep{li2024snapkv}.
As shown in Table~\ref{tab:snapkv}, \name~does not exhibit significant performance degradation. 
We left it for future work to combine \name~with existing mixed-precision quantization for further memory saving~\citep{quarot, dong2024qaq, yang2024tokenleftbehindreliable}.

\begin{table*}[htbp]
\caption{LongBench Evaluations of \name\ with SnapKV. }
\centering
\label{tab:snapkv}
\vskip 0.1in
\resizebox{0.88\textwidth}{!}{
    
    \begin{tabular}{cccccccccc}
    \toprule
    ${\textbf{\textrm{LLM}}}$ & \textbf{NtrvQA} & \textbf{Qasper} & \textbf{MF-en} & \textbf{2Wiki} & \textbf{Hotpot} & \textbf{Musique} & \textbf{Lcc} & \textbf{RepoBench} & \textbf{Avg.}\\
    \midrule
    \textbf{\textrm{\textit{Full KV}}} & 31.36 & 46.78 & 56.72 & 49.47 & 58.54 & 32.23 & 63.28 & 56.73 & \textbf{49.39} \\
    \midrule
    \textbf{\textrm{\textit{SnapKV}}: 4096}~~\quad & 31.31 & 46.51 & 57.03 & 49.71 & 57.99 & 32.79 & 62.90 & 55.75 & 49.25~(-0.14) \\
    \cmidrule[0.5pt](lr{1pt}){1-1}
    \multicolumn{1}{r}{\textit{\textbf{w.}}~\textbf{\textit{\name}}} & 31.21 & 46.38 & 56.45 & 49.38 & 58.05 & 31.99 & 62.90 & 55.75 & 49.01~(-0.38) \\
    \midrule
    \textbf{\textrm{\textit{SnapKV}}: 1024}~~\quad & 31.51 & 43.45 & 56.14 & 49.61 & 57.78 & 32.01 & 62.45 & 56.00 & 48.62~(-0.77) \\
    \cmidrule[0.5pt](lr{1pt}){1-1}
    \multicolumn{1}{r}{\textit{\textbf{w.}}~\textbf{\textit{\name}}} & 31.27 & 42.61 & 55.18 & 48.60 & 57.49 & 31.12 & 61.54 & 54.57 & 47.80~(-1.59) \\
    \bottomrule
    \end{tabular}
}
\end{table*}

% \begin{table*}[h]
% \caption{\textcolor{red}{Caption!!!}}
% \centering
% \label{tab:snapkv}
% \vskip 0.1in
% \resizebox{0.95\textwidth}{!}{
    
%     \begin{tabular}{cccccccccc}
%     \toprule
%     ${\textbf{\textrm{LLM}}}^{\mathbf{*}}$ & \textbf{NtrvQA} & \textbf{Qasper} & \textbf{MF-en} & \textbf{2Wiki} & \textbf{Hotpot} & \textbf{Musique} & \textbf{Lcc} & \textbf{RepoBench} & \textbf{Avg.~$\uparrow$}\\
%     \midrule
%     \textbf{\textrm{\textit{Full KV}}} & 31.36 & 46.78 & 56.72 & 49.47 & 58.54 & 32.23 & 63.28 & 56.73 & \cellcolor{blue!10}\textbf{49.39} \\
%     \midrule
%     \textbf{\textrm{\textit{SnapKV}}: 4096}~~\quad & 31.31 & 46.51 & 57.03 & 49.71 & 57.99 & 32.79 & 62.90 & 55.75 & 49.25~(-0.14) \\
%     \cmidrule[0.5pt](lr{3pt}){1-1}
%     \multicolumn{1}{r|}{\textit{\textbf{w.}}~\textbf{\textit{\name}}} & 31.21 & 46.38 & 56.45 & 49.38 & 58.05 & 31.99 & 62.90 & 55.75 & 49.01~(-0.38) \\
%     \midrule
%     \textbf{\textrm{\textit{SnapKV}}: 1024}~~\quad & 31.51 & 43.45 & 56.14 & 49.61 & 57.78 & 32.01 & 62.45 & 56.00 & 48.62~(-0.77) \\
%     \cmidrule[0.5pt](lr{3pt}){1-1}
%     \multicolumn{1}{r|}{\textit{\textbf{w.}}~\textbf{\textit{\name}}} & 31.27 & 42.61 & 55.18 & 48.60 & 57.49 & 31.12 & 61.54 & 54.57 & 47.80~(-1.59) \\
%     \bottomrule
%     \end{tabular}
% }
% \end{table*}

% \subsection{Performance of \name~under low bits setting.}

\section{Conclusion}

In this paper, we view the outliers in the key cache of LLMs from a novel polar-coordinate-based perspective, which provides an efficient and effective solution, \name, to reduce the complexity and quantization costs in previous methods.
\name\ well preserves downstream performance even in long-context understanding and long chain-of-thought generation, comparable to previous works under 4-bit precision while achieving superior efficiency.
We hope the polar coordinate view can inspire the community to advance new low-bit precision quantization techniques.

\newpage

%%%%%%%%%%%%%%%%%%%%%%%%%%%%%%%%%%%%%%%%%%%%%%%%%%%%%%%%%%%%

\section*{Acknowledgments}
This work is supported by Meituan through Agentic system X Program. 
Songhao Wu is supported by "Qiushi Academic-Dongliang" Project of Renmin
University of China (No. RUC24QSDL015). 
We appreciate all reviewers for their valuable comments and suggestions, which are crucial for improving our work. 
We are also grateful to Jixiang Hong, Tao Tan, Jia-Nan Li and Yuxuan Liu for their insightful suggestions on the manuscript.

\bibliographystyle{plainnat}
\bibliography{neurips_2025_refs}

@misc{o12024,
	author = {OpenAI},
	title = {Learning to Reason with LLMs
},
	howpublished = {\url{https://openai.com/index/learning-to-reason-with-llms/}},
	year = {2024},
	note = {[Accessed 19-09-2024]},
}

@article{liu-etal-2024-lost,
    title = "Lost in the Middle: How Language Models Use Long Contexts",
    author = "Liu, Nelson F.  and
      Lin, Kevin  and
      Hewitt, John  and
      Paranjape, Ashwin  and
      Bevilacqua, Michele  and
      Petroni, Fabio  and
      Liang, Percy",
    journal = "Transactions of the Association for Computational Linguistics",
    volume = "12",
    year = "2024",
    address = "Cambridge, MA",
    publisher = "MIT Press",
    url = "https://aclanthology.org/2024.tacl-1.9/",
    doi = "10.1162/tacl_a_00638",
    pages = "157--173"
}

@misc{bahdanau2016neuralmachinetranslationjointly,
      title={Neural Machine Translation by Jointly Learning to Align and Translate}, 
      author={Dzmitry Bahdanau and Kyunghyun Cho and Yoshua Bengio},
      year={2016},
      eprint={1409.0473},
      archivePrefix={arXiv},
      primaryClass={cs.CL},
      url={https://arxiv.org/abs/1409.0473}, 
}

@inproceedings{transformer,
 author = {Vaswani, Ashish and Shazeer, Noam and Parmar, Niki and Uszkoreit, Jakob and Jones, Llion and Gomez, Aidan N and Kaiser, \L ukasz and Polosukhin, Illia},
 booktitle = {Advances in Neural Information Processing Systems},
 editor = {I. Guyon and U. Von Luxburg and S. Bengio and H. Wallach and R. Fergus and S. Vishwanathan and R. Garnett},
 pages = {},
 publisher = {Curran Associates, Inc.},
 title = {Attention is All you Need},
 url = {https://proceedings.neurips.cc/paper_files/paper/2017/file/3f5ee243547dee91fbd053c1c4a845aa-Paper.pdf},
 volume = {30},
 year = {2017}
}

@misc{deepseekai2024deepseekv2strongeconomicalefficient,
      title={DeepSeek-V2: A Strong, Economical, and Efficient Mixture-of-Experts Language Model}, 
      author={DeepSeek-AI and Aixin Liu and Bei Feng and Bin Wang and Bingxuan Wang and Bo Liu and Chenggang Zhao and Chengqi Dengr and Chong Ruan and Damai Dai and Daya Guo and Dejian Yang and Deli Chen and Dongjie Ji and Erhang Li and Fangyun Lin and Fuli Luo and Guangbo Hao and Guanting Chen and Guowei Li and H. Zhang and Hanwei Xu and Hao Yang and Haowei Zhang and Honghui Ding and Huajian Xin and Huazuo Gao and Hui Li and Hui Qu and J. L. Cai and Jian Liang and Jianzhong Guo and Jiaqi Ni and Jiashi Li and Jin Chen and Jingyang Yuan and Junjie Qiu and Junxiao Song and Kai Dong and Kaige Gao and Kang Guan and Lean Wang and Lecong Zhang and Lei Xu and Leyi Xia and Liang Zhao and Liyue Zhang and Meng Li and Miaojun Wang and Mingchuan Zhang and Minghua Zhang and Minghui Tang and Mingming Li and Ning Tian and Panpan Huang and Peiyi Wang and Peng Zhang and Qihao Zhu and Qinyu Chen and Qiushi Du and R. J. Chen and R. L. Jin and Ruiqi Ge and Ruizhe Pan and Runxin Xu and Ruyi Chen and S. S. Li and Shanghao Lu and Shangyan Zhou and Shanhuang Chen and Shaoqing Wu and Shengfeng Ye and Shirong Ma and Shiyu Wang and Shuang Zhou and Shuiping Yu and Shunfeng Zhou and Size Zheng and T. Wang and Tian Pei and Tian Yuan and Tianyu Sun and W. L. Xiao and Wangding Zeng and Wei An and Wen Liu and Wenfeng Liang and Wenjun Gao and Wentao Zhang and X. Q. Li and Xiangyue Jin and Xianzu Wang and Xiao Bi and Xiaodong Liu and Xiaohan Wang and Xiaojin Shen and Xiaokang Chen and Xiaosha Chen and Xiaotao Nie and Xiaowen Sun and Xiaoxiang Wang and Xin Liu and Xin Xie and Xingkai Yu and Xinnan Song and Xinyi Zhou and Xinyu Yang and Xuan Lu and Xuecheng Su and Y. Wu and Y. K. Li and Y. X. Wei and Y. X. Zhu and Yanhong Xu and Yanping Huang and Yao Li and Yao Zhao and Yaofeng Sun and Yaohui Li and Yaohui Wang and Yi Zheng and Yichao Zhang and Yiliang Xiong and Yilong Zhao and Ying He and Ying Tang and Yishi Piao and Yixin Dong and Yixuan Tan and Yiyuan Liu and Yongji Wang and Yongqiang Guo and Yuchen Zhu and Yuduan Wang and Yuheng Zou and Yukun Zha and Yunxian Ma and Yuting Yan and Yuxiang You and Yuxuan Liu and Z. Z. Ren and Zehui Ren and Zhangli Sha and Zhe Fu and Zhen Huang and Zhen Zhang and Zhenda Xie and Zhewen Hao and Zhihong Shao and Zhiniu Wen and Zhipeng Xu and Zhongyu Zhang and Zhuoshu Li and Zihan Wang and Zihui Gu and Zilin Li and Ziwei Xie},
      year={2024},
      eprint={2405.04434},
      archivePrefix={arXiv},
      primaryClass={cs.CL},
      url={https://arxiv.org/abs/2405.04434}, 
}

@inproceedings{ainslie-etal-2023-gqa,
    title = "{GQA}: Training Generalized Multi-Query Transformer Models from Multi-Head Checkpoints",
    author = "Ainslie, Joshua  and
      Lee-Thorp, James  and
      de Jong, Michiel  and
      Zemlyanskiy, Yury  and
      Lebron, Federico  and
      Sanghai, Sumit",
    editor = "Bouamor, Houda  and
      Pino, Juan  and
      Bali, Kalika",
    booktitle = "Proceedings of the 2023 Conference on Empirical Methods in Natural Language Processing",
    month = dec,
    year = "2023",
    address = "Singapore",
    publisher = "Association for Computational Linguistics",
    url = "https://aclanthology.org/2023.emnlp-main.298/",
    doi = "10.18653/v1/2023.emnlp-main.298",
    pages = "4895--4901"
}

@article{li2024snapkv,
  title={Snapkv: Llm knows what you are looking for before generation},
  author={Li, Yuhong and Huang, Yingbing and Yang, Bowen and Venkitesh, Bharat and Locatelli, Acyr and Ye, Hanchen and Cai, Tianle and Lewis, Patrick and Chen, Deming},
  journal={arXiv preprint arXiv:2404.14469},
  year={2024}
}

@misc{zhang2023h2oheavyhitteroracleefficient,
      title={H$_2$O: Heavy-Hitter Oracle for Efficient Generative Inference of Large Language Models}, 
      author={Zhenyu Zhang and Ying Sheng and Tianyi Zhou and Tianlong Chen and Lianmin Zheng and Ruisi Cai and Zhao Song and Yuandong Tian and Christopher Ré and Clark Barrett and Zhangyang Wang and Beidi Chen},
      year={2023},
      eprint={2306.14048},
      archivePrefix={arXiv},
      primaryClass={cs.LG},
      url={https://arxiv.org/abs/2306.14048}, 
}

@misc{cai2024pyramidkvdynamickvcache,
      title={PyramidKV: Dynamic KV Cache Compression based on Pyramidal Information Funneling}, 
      author={Zefan Cai and Yichi Zhang and Bofei Gao and Yuliang Liu and Tianyu Liu and Keming Lu and Wayne Xiong and Yue Dong and Baobao Chang and Junjie Hu and Wen Xiao},
      year={2024},
      eprint={2406.02069},
      archivePrefix={arXiv},
      primaryClass={cs.CL},
      url={https://arxiv.org/abs/2406.02069}, 
}

@inproceedings{kivi,
author = {Liu, Zirui and Yuan, Jiayi and Jin, Hongye and Zhong, Shaochen (Henry) and Xu, Zhaozhuo and Braverman, Vladimir and Chen, Beidi and Hu, Xia},
title = {KIVI: a tuning-free asymmetric 2bit quantization for KV cache},
year = {2025},
publisher = {JMLR.org},
booktitle = {Proceedings of the 41st International Conference on Machine Learning},
articleno = {1311},
numpages = {13},
location = {Vienna, Austria},
series = {ICML'24}
}

@misc{kvquant,
      title={KVQuant: Towards 10 Million Context Length LLM Inference with KV Cache Quantization}, 
      author={Coleman Hooper and Sehoon Kim and Hiva Mohammadzadeh and Michael W. Mahoney and Yakun Sophia Shao and Kurt Keutzer and Amir Gholami},
      year={2024},
      eprint={2401.18079},
      archivePrefix={arXiv},
      primaryClass={cs.LG},
      url={https://arxiv.org/abs/2401.18079}, 
}

@misc{zipcache,
      title={ZipCache: Accurate and Efficient KV Cache Quantization with Salient Token Identification}, 
      author={Yefei He and Luoming Zhang and Weijia Wu and Jing Liu and Hong Zhou and Bohan Zhuang},
      year={2024},
      eprint={2405.14256},
      archivePrefix={arXiv},
      primaryClass={cs.LG},
      url={https://arxiv.org/abs/2405.14256}, 
}

@inproceedings{atom,
 author = {Zhao, Yilong and Lin, Chien-Yu and Zhu, Kan and Ye, Zihao and Chen, Lequn and Zheng, Size and Ceze, Luis and Krishnamurthy, Arvind and Chen, Tianqi and Kasikci, Baris},
 booktitle = {Proceedings of Machine Learning and Systems},
 editor = {P. Gibbons and G. Pekhimenko and C. De Sa},
 pages = {196--209},
 title = {Atom: Low-Bit Quantization for Efficient and Accurate LLM Serving},
 url = {https://proceedings.mlsys.org/paper_files/paper/2024/file/5edb57c05c81d04beb716ef1d542fe9e-Paper-Conference.pdf},
 volume = {6},
 year = {2024}
}

@misc{gear,
      title={GEAR: An Efficient KV Cache Compression Recipe for Near-Lossless Generative Inference of LLM}, 
      author={Hao Kang and Qingru Zhang and Souvik Kundu and Geonhwa Jeong and Zaoxing Liu and Tushar Krishna and Tuo Zhao},
      year={2024},
      eprint={2403.05527},
      archivePrefix={arXiv},
      primaryClass={cs.LG},
      url={https://arxiv.org/abs/2403.05527}, 
}

@misc{rope,
      title={RoFormer: Enhanced Transformer with Rotary Position Embedding}, 
      author={Jianlin Su and Yu Lu and Shengfeng Pan and Ahmed Murtadha and Bo Wen and Yunfeng Liu},
      year={2023},
      eprint={2104.09864},
      archivePrefix={arXiv},
      primaryClass={cs.CL},
      url={https://arxiv.org/abs/2104.09864}, 
}

@misc{touvron2023llama2,
      title={Llama 2: Open Foundation and Fine-Tuned Chat Models}, 
      author={Hugo Touvron and Louis Martin and Kevin Stone and Peter Albert and Amjad Almahairi and Yasmine Babaei and Nikolay Bashlykov and Soumya Batra and Prajjwal Bhargava and Shruti Bhosale and Dan Bikel and Lukas Blecher and Cristian Canton Ferrer and Moya Chen and Guillem Cucurull and David Esiobu and Jude Fernandes and Jeremy Fu and Wenyin Fu and Brian Fuller and Cynthia Gao and Vedanuj Goswami and Naman Goyal and Anthony Hartshorn and Saghar Hosseini and Rui Hou and Hakan Inan and Marcin Kardas and Viktor Kerkez and Madian Khabsa and Isabel Kloumann and Artem Korenev and Punit Singh Koura and Marie-Anne Lachaux and Thibaut Lavril and Jenya Lee and Diana Liskovich and Yinghai Lu and Yuning Mao and Xavier Martinet and Todor Mihaylov and Pushkar Mishra and Igor Molybog and Yixin Nie and Andrew Poulton and Jeremy Reizenstein and Rashi Rungta and Kalyan Saladi and Alan Schelten and Ruan Silva and Eric Michael Smith and Ranjan Subramanian and Xiaoqing Ellen Tan and Binh Tang and Ross Taylor and Adina Williams and Jian Xiang Kuan and Puxin Xu and Zheng Yan and Iliyan Zarov and Yuchen Zhang and Angela Fan and Melanie Kambadur and Sharan Narang and Aurelien Rodriguez and Robert Stojnic and Sergey Edunov and Thomas Scialom},
      year={2023},
      eprint={2307.09288},
      archivePrefix={arXiv},
      primaryClass={cs.CL},
      url={https://arxiv.org/abs/2307.09288}, 
}

@article{llama3,
title={Llama 3 Model Card},
author={AI@Meta},
year={2024},
url = {https://github.com/meta-llama/llama3/blob/main/MODEL_CARD.md}
}

@misc{qwen2025qwen25,
    title = {Qwen2.5: A Party of Foundation Models},
    url = {https://qwenlm.github.io/blog/qwen2.5/},
    author = {Qwen Team},
    month = {September},
    year = {2024}
}

@inproceedings{wolf-etal-2020-transformers,
    title = "Transformers: State-of-the-Art Natural Language Processing",
    author = "Thomas Wolf and Lysandre Debut and Victor Sanh and Julien Chaumond and Clement Delangue and Anthony Moi and Pierric Cistac and Tim Rault and Rémi Louf and Morgan Funtowicz and Joe Davison and Sam Shleifer and Patrick von Platen and Clara Ma and Yacine Jernite and Julien Plu and Canwen Xu and Teven Le Scao and Sylvain Gugger and Mariama Drame and Quentin Lhoest and Alexander M. Rush",
    booktitle = "Proceedings of the 2020 Conference on Empirical Methods in Natural Language Processing: System Demonstrations",
    month = oct,
    year = "2020",
    address = "Online",
    publisher = "Association for Computational Linguistics",
    url = "https://www.aclweb.org/anthology/2020.emnlp-demos.6",
    pages = "38--45"
}

@article{bai2023longbench,
  title={LongBench: A Bilingual, Multitask Benchmark for Long Context Understanding},
  author={Bai, Yushi and Lv, Xin and Zhang, Jiajie and Lyu, Hongchang and Tang, Jiankai and Huang, Zhidian and Du, Zhengxiao and Liu, Xiao and Zeng, Aohan and Hou, Lei and Dong, Yuxiao and Tang, Jie and Li, Juanzi},
  journal={arXiv preprint arXiv:2308.14508},
  year={2023}
}

@article{cobbe2021gsm8k,
  title={Training Verifiers to Solve Math Word Problems},
  author={Cobbe, Karl and Kosaraju, Vineet and Bavarian, Mohammad and Chen, Mark and Jun, Heewoo and Kaiser, Lukasz and Plappert, Matthias and Tworek, Jerry and Hilton, Jacob and Nakano, Reiichiro and Hesse, Christopher and Schulman, John},
  journal={arXiv preprint arXiv:2110.14168},
  year={2021}
}

@misc{lightman2023math500,
      title={Let's Verify Step by Step}, 
      author={Hunter Lightman and Vineet Kosaraju and Yura Burda and Harri Edwards and Bowen Baker and Teddy Lee and Jan Leike and John Schulman and Ilya Sutskever and Karl Cobbe},
      year={2023},
      eprint={2305.20050},
      archivePrefix={arXiv},
      primaryClass={cs.LG},
      url={https://arxiv.org/abs/2305.20050}, 
}

@misc{rein2023gpqa,
      title={GPQA: A Graduate-Level Google-Proof Q\&A Benchmark}, 
      author={David Rein and Betty Li Hou and Asa Cooper Stickland and Jackson Petty and Richard Yuanzhe Pang and Julien Dirani and Julian Michael and Samuel R. Bowman},
      year={2023},
      eprint={2311.12022},
      archivePrefix={arXiv},
      primaryClass={cs.AI},
      url={https://arxiv.org/abs/2311.12022}, 
}

@misc{liu2025hurts,
      title={Quantization Hurts Reasoning? An Empirical Study on Quantized Reasoning Models}, 
      author={Ruikang Liu and Yuxuan Sun and Manyi Zhang and Haoli Bai and Xianzhi Yu and Tiezheng Yu and Chun Yuan and Lu Hou},
      year={2025},
      eprint={2504.04823},
      archivePrefix={arXiv},
      primaryClass={cs.CL},
      url={https://arxiv.org/abs/2504.04823}, 
}

@misc{luo2024r1,
      title={Improve Mathematical Reasoning in Language Models by Automated Process Supervision}, 
      author={Liangchen Luo and Yinxiao Liu and Rosanne Liu and Samrat Phatale and Meiqi Guo and Harsh Lara and Yunxuan Li and Lei Shu and Yun Zhu and Lei Meng and Jiao Sun and Abhinav Rastogi},
      year={2024},
      eprint={2406.06592},
      archivePrefix={arXiv},
      primaryClass={cs.CL},
      url={https://arxiv.org/abs/2406.06592}, 
}

@misc{zandieh2024qjl,
      title={QJL: 1-Bit Quantized JL Transform for KV Cache Quantization with Zero Overhead}, 
      author={Amir Zandieh and Majid Daliri and Insu Han},
      year={2024},
      eprint={2406.03482},
      archivePrefix={arXiv},
      primaryClass={cs.LG},
      url={https://arxiv.org/abs/2406.03482}, 
}

@misc{wei2023chainofthought,
      title={Chain-of-Thought Prompting Elicits Reasoning in Large Language Models}, 
      author={Jason Wei and Xuezhi Wang and Dale Schuurmans and Maarten Bosma and Brian Ichter and Fei Xia and Ed Chi and Quoc Le and Denny Zhou},
      year={2023},
      eprint={2201.11903},
      archivePrefix={arXiv},
      primaryClass={cs.CL},
      url={https://arxiv.org/abs/2201.11903}, 
}

@misc{brown2020gpt3,
      title={Language Models are Few-Shot Learners}, 
      author={Tom B. Brown and Benjamin Mann and Nick Ryder and Melanie Subbiah and Jared Kaplan and Prafulla Dhariwal and Arvind Neelakantan and Pranav Shyam and Girish Sastry and Amanda Askell and Sandhini Agarwal and Ariel Herbert-Voss and Gretchen Krueger and Tom Henighan and Rewon Child and Aditya Ramesh and Daniel M. Ziegler and Jeffrey Wu and Clemens Winter and Christopher Hesse and Mark Chen and Eric Sigler and Mateusz Litwin and Scott Gray and Benjamin Chess and Jack Clark and Christopher Berner and Sam McCandlish and Alec Radford and Ilya Sutskever and Dario Amodei},
      year={2020},
      eprint={2005.14165},
      archivePrefix={arXiv},
      primaryClass={cs.CL},
      url={https://arxiv.org/abs/2005.14165}, 
}

@misc{shazeer2019mqa,
      title={Fast Transformer Decoding: One Write-Head is All You Need}, 
      author={Noam Shazeer},
      year={2019},
      eprint={1911.02150},
      archivePrefix={arXiv},
      primaryClass={cs.NE},
      url={https://arxiv.org/abs/1911.02150}, 
}

@misc{yang2024tokenleftbehindreliable,
      title={No Token Left Behind: Reliable KV Cache Compression via Importance-Aware Mixed Precision Quantization}, 
      author={June Yong Yang and Byeongwook Kim and Jeongin Bae and Beomseok Kwon and Gunho Park and Eunho Yang and Se Jung Kwon and Dongsoo Lee},
      year={2024},
      eprint={2402.18096},
      archivePrefix={arXiv},
      primaryClass={cs.LG},
      url={https://arxiv.org/abs/2402.18096}, 
}

@misc{dong2024qaq,
      title={QAQ: Quality Adaptive Quantization for LLM KV Cache}, 
      author={Shichen Dong and Wen Cheng and Jiayu Qin and Wei Wang},
      year={2024},
      eprint={2403.04643},
      archivePrefix={arXiv},
      primaryClass={cs.CL},
      url={https://arxiv.org/abs/2403.04643}, 
}

@misc{lighteval,
  author = {Habib, Nathan and Fourrier, Clémentine and Kydlíček, Hynek and Wolf, Thomas and Tunstall, Lewis},
  title = {LightEval: A lightweight framework for LLM evaluation},
  year = {2023},
  version = {0.8.0},
  url = {https://github.com/huggingface/lighteval}
}

@article{pqcache,
  title={Pqcache: Product quantization-based kvcache for long context llm inference},
  author={Zhang, Hailin and Ji, Xiaodong and Chen, Yilin and Fu, Fangcheng and Miao, Xupeng and Nie, Xiaonan and Chen, Weipeng and Cui, Bin},
  journal={Proceedings of the ACM on Management of Data},
  volume={3},
  number={3},
  pages={1--30},
  year={2025},
  publisher={ACM New York, NY, USA}
}

@misc{codellama,
      title={Code Llama: Open Foundation Models for Code}, 
      author={Baptiste Rozière and Jonas Gehring and Fabian Gloeckle and Sten Sootla and Itai Gat and Xiaoqing Ellen Tan and Yossi Adi and Jingyu Liu and Romain Sauvestre and Tal Remez and Jérémy Rapin and Artyom Kozhevnikov and Ivan Evtimov and Joanna Bitton and Manish Bhatt and Cristian Canton Ferrer and Aaron Grattafiori and Wenhan Xiong and Alexandre Défossez and Jade Copet and Faisal Azhar and Hugo Touvron and Louis Martin and Nicolas Usunier and Thomas Scialom and Gabriel Synnaeve},
      year={2024},
      eprint={2308.12950},
      archivePrefix={arXiv},
      primaryClass={cs.CL},
      url={https://arxiv.org/abs/2308.12950}, 
}

@misc{pytorch,
      title={PyTorch: An Imperative Style, High-Performance Deep Learning Library}, 
      author={Adam Paszke and Sam Gross and Francisco Massa and Adam Lerer and James Bradbury and Gregory Chanan and Trevor Killeen and Zeming Lin and Natalia Gimelshein and Luca Antiga and Alban Desmaison and Andreas Köpf and Edward Yang and Zach DeVito and Martin Raison and Alykhan Tejani and Sasank Chilamkurthy and Benoit Steiner and Lu Fang and Junjie Bai and Soumith Chintala},
      year={2019},
      eprint={1912.01703},
      archivePrefix={arXiv},
      primaryClass={cs.LG},
      url={https://arxiv.org/abs/1912.01703}, 
}

@misc{qwen3,
      title={Qwen3 Technical Report}, 
      author={An Yang and Anfeng Li and Baosong Yang and Beichen Zhang and Binyuan Hui and Bo Zheng and Bowen Yu and Chang Gao and Chengen Huang and Chenxu Lv and Chujie Zheng and Dayiheng Liu and Fan Zhou and Fei Huang and Feng Hu and Hao Ge and Haoran Wei and Huan Lin and Jialong Tang and Jian Yang and Jianhong Tu and Jianwei Zhang and Jianxin Yang and Jiaxi Yang and Jing Zhou and Jingren Zhou and Junyang Lin and Kai Dang and Keqin Bao and Kexin Yang and Le Yu and Lianghao Deng and Mei Li and Mingfeng Xue and Mingze Li and Pei Zhang and Peng Wang and Qin Zhu and Rui Men and Ruize Gao and Shixuan Liu and Shuang Luo and Tianhao Li and Tianyi Tang and Wenbiao Yin and Xingzhang Ren and Xinyu Wang and Xinyu Zhang and Xuancheng Ren and Yang Fan and Yang Su and Yichang Zhang and Yinger Zhang and Yu Wan and Yuqiong Liu and Zekun Wang and Zeyu Cui and Zhenru Zhang and Zhipeng Zhou and Zihan Qiu},
      year={2025},
      eprint={2505.09388},
      archivePrefix={arXiv},
      primaryClass={cs.CL},
      url={https://arxiv.org/abs/2505.09388}, 
}

@misc{quarot,
      title={QuaRot: Outlier-Free 4-Bit Inference in Rotated LLMs}, 
      author={Saleh Ashkboos and Amirkeivan Mohtashami and Maximilian L. Croci and Bo Li and Pashmina Cameron and Martin Jaggi and Dan Alistarh and Torsten Hoefler and James Hensman},
      year={2024},
      eprint={2404.00456},
      archivePrefix={arXiv},
      primaryClass={cs.LG},
      url={https://arxiv.org/abs/2404.00456}, 
}

\newpage

\newpage

\appendix

\section{Accelerated Query-Key Inner Product with Cached Keys in \name}
\label{sec:algorithm}
For a quantized representation \(\Big( Q(\rho_{n}[j]), Q(\theta_{n}[j]) \Big)\), we first dequantize the polar coordinates back into Cartesian coordinates, which serves as  the at dimensions \( 2j \) and \( 2j+1 \):
\[
\begin{bmatrix}
 \bm{\widetilde{{\mathbf{K}}}}_n[2j] \\
 \bm{\widetilde{{\mathbf{K}}}}_n[2j+1]
\end{bmatrix}^{\top}
=
\begin{bmatrix}
\tilde{\rho}_{n}[j] \cdot \cos \left( \tilde{\theta}_{n}[j] \right) \\
\tilde{\rho}_{n}[j] \cdot \sin \left( \tilde{\theta}_{n}[j] \right) 
\end{bmatrix}^{\top},
\]
where \( \tilde{\rho}_{n}[j],~\tilde{\theta}_{n}[j]\) is the dequantization for \( \left(\rho,\theta\right) \) respectively. 
\(\tilde{\rho}_{n}[j]\) and \(\tilde{\theta}_{n}[j])\) are formulated as:
\[
\tilde{\rho}_{n}[j] = \big(Q(\rho_{n}[j]) + \frac{1}{2} \big) \cdot s_{\rho[:]}{[j]} + z_{\rho[:]}{[j]}, \quad \tilde{\theta}_{n}[j] = \big(Q(\theta_{n}[j]) + \frac{1}{2} \big) \cdot s_{\theta[:]}{[j]} + z_{\theta[:]}{[j]}.
\]
\(\Big( Q(\rho_{n}[j]), Q(\theta_{n}[j]) \Big)\) represents \(
\begin{bmatrix}
\widetilde{\mathbf{K}}_{n}[2j],\;
\widetilde{\mathbf{K}}_{n}[2j+1]
\end{bmatrix}
\) as a state in the lookup table.

When computing the query-key inner product for the dimensions \( 2j \) and \( 2j+1 \), the result is:
\[
\mathbf{product}_{2j, 2j+1} = \mathbf{Q}_{n}[2j] \cdot \widetilde{\mathbf{K}}_{n}[2j] + \mathbf{Q}_{n}[2j+1] \cdot \widetilde{\mathbf{K}}_{t}[2j+1],
\]
and the final inner product is the sum:
\[
\sum_{0 \leq j < \frac{d}{2}} \mathbf{product}_{2j, 2j+1}.
\]

Figure~\ref{fig:algorithm} presents a naive PyTorch~\citep{pytorch} implementation of the aforementioned dequantization-and-multiplication operation. 
We implement a fused Triton kernel that reproduces the functionality of the PyTorch code for improved computational efficiency.
More details can be found in our code.

\begin{figure*}[!hbtp]
\begin{lstlisting}[language={Python}]
import torch


def attention_decode_forward_pytorch_impl(
    q,  # q's shape: (B, N, 1, 2, D)
    r, rscale, rmn,  # r's shape: (B, N, G, D)
    t, tscale, tmn,  # t's shape: (B, N, G, D)
    # rscale, rmn, tscale and tmn's shape: (B, N, 1, 1, D)
    rbits: int = 4, tbits: int = 4,
):
    # phi: finite set for theta
    phi = torch.arange(0, 2 ** tbits)[None, None, :, None, None]
    phi = (2 * phi + 1) / 2 * tscale + tmn

    # rho: finite set for rho
    rho = torch.arange(0, 2 ** rbits)[None, None, :, None, None]
    rho = (2 * rho + 1) / 2 * rscale + rmn  

    phi = torch.cat([phi.cos(), phi.sin()], dim=-2)  
   
    accumulator = torch.sum(q * phi, dim=-2)  # (B, N, 2^tbits, D)
    accumulator = torch.gather(accumulator, 2, t.unsqueeze(-1).expand_as(accumulator))
    accumulator *= torch.gather(rho.squeeze(-2), 2, r.unsqueeze(-1).expand_as(accumulator))

    accumulator = accumulator.sum(-1)

    return accumulator
\end{lstlisting} 
\caption{Pytorch implementation of the accelerated Query-Key Inner Product in \name. }
\label{fig:algorithm}
\end{figure*}

\newpage

\section{Details experiment setup}
\label{sec:main_setup}
In this section, we provide additional details about the experimental setups.

\subsection{Setup of baseline methods}
This section provides an overview of the baseline methods.
Following this, we outline the quantization configurations and variants for both the baselines and our proposed \name.
We also explain how the average number of bits for the quantization parameters is calculated.

% \begin{itemize}
%     \item \textit{\textbf{ZipCache}~\citep{zipcache}}~proposes a channel-separable token-wise scheme for key quantization.
%     \item \textit{\textbf{KIVI}~\citep{kivi}}~introduces a group-wise activation quantization, which adopts channel-wise quantization for key cache and token-wise quantization for value. 
%     \item \textit{\textbf{QJL}~\citep{zandieh2024qjl}} applies Johnson-Lindenstrauss transformation to key embeddings and eliminates memory overheads for storing  quantization constants.
% \end{itemize}

\textrm{\textbf{Int-}$\mathbf{N}$} applies token-wise \textrm{N}-bit quantization to the key states. 
This token-wise quantization incurs $32/d$ bits quantization parameters~(16 bits for zero-points and 16 bits for scaling factors per token), which amounts to 0.25 bits per token when $d=128$.
    
\textrm{\textbf{ZipCache}-$\mathbf{N}$}~\citep{zipcache} introduces a channel-separable, token-wise scheme for key quantization, where $N$ denotes the quantization bits.
Each key channel is normalized by the square root of its maximum magnitude before quantization.
Similar to \textrm{\textbf{Int}-$\mathbf{N}$}, this method performs token-wise quantization, allocating 0.25 bits for zero-points and scales.
    
\textrm{\textbf{KIVI}-$\mathbf{N}$}~\citep{kivi} employs an asymmetric strategy for KV cache quantization, applying channel-wise quantization to the key cache and token-wise quantization to the value cache.
For 4-bit quantization, we use \textrm{\textbf{KIVI}-$\mathrm{4}$} with a group size of 128.
For 3-bit quantization, we use \textrm{\textbf{KIVI}-$\mathrm{2}$} with a group size of 32, as the official implementation does not support 3-bit quantization.
This channel-wise quantization introduces $(16+16)d$ bits of quantization parameters per group, which increases the average bitwidth by $32/g$:
1 bit for $g=32$ and 0.25 bits for $g=128$ respectively.

\textrm{\textbf{QJL}}~\citep{zandieh2024qjl} applies Johnson-Lindenstrauss transformation to key states, removing the memory overheads associated with storing quantization constants.
For a 3-bit quantization schema, this method achieves a key cache bitwidth of 3.13 bits.

${\textbf{\textrm{\name}}_{rt}}$ assigns \textrm{r} bits for radii quantization and \textrm{t} bits for polar angles, resulting in $(r+t)/2$ bitwidth for key states quantization. 
\name\ also employs channel-wise quantization with grouping;
this group partitioning adds an overhead of $32/g$ bits.

Nearly all methods discussed here require a buffer size or residual length for applying quantization.
We exclude the contribution of these residual key states to bit counts for comparisons.

\subsection{Configurations of the open-sourced LLMs}
\textbullet \quad \textbf{\textrm{Qwen-2.5-1.5B-Instruct}}~\citep{qwen2025qwen25} is the instruction-tuned 1.5B Qwen2.5 model, which supports a context length of up to 131,072 tokens and features a base RoPE frequency of 1,000,000.

\textbullet \quad \textbf{\textrm{Llama-2-7B-Chat}}~\citep{touvron2023llama2} has a context length of 4096 and base RoPE frequency of 10,000.

\textbullet \quad \textbf{\textrm{Llama-3.1-8B-Instruct}}~\citep{llama3} is an 8 billion parameter language model, designed to handle a context length of up to 131,072 tokens, and has the base RoPE frequency set to 500,000.

\subsection{Reasoning model evaluation}

For fair comparison, the evaluation code is built on the Huggingface Lighteval framework~\citep{lighteval}. 
We use the default generation configuration for datasets splitting, and the EM score is reported in Table~\ref{tab:reasoning}. 
More implementation details can be found in our released code.

\section{NTK RoPE scaling experiment}
\label{sec:rope}
We adopt NTK RoPE scaling~\citep{codellama} to extend the context window of the Llama-2-7B-Chat model from $4096$ to $8192$. 
Specifically, we implement dynamic RoPE updates based on the Hugging Face codebase.
\name achieve an average performance of 32.44 on LongBench.
Compared to the 32.15 score of the Bf16 baseline, \name's performance remains competitive.

\section{Sensitivity Analysis of Key-Value Quantization}
\label{sec:apdx_sensitive}
In Section~\ref{sec:sensitive}, we combine 4-bit key quantization $\textrm{PolarQuant}_{44}$ 
with 2-bit value quantization and observe minimal performance degradation.
We further evaluate value quantization (KIVI, group size 128) on LongBench.
By retaining the key cache at full precision, KIVI result in no performance drop. 
However, when the key is quantized to the same bitwidth while the value is kept at full precision, the performance drops significantly. 
Table~\ref{tab:sensitive} presents the results, the notation $\mathbf{(K_b, V_c)}$ denotes key quantization to $b$-bit and value quantization to $c$-bit.

\begin{table*}[htbp]
    \centering
    \small
    \caption{Impact of key and value quantization bitwidths on LongBench performance. }
    \vskip 0.1in
    \label{tab:sensitive}
    \resizebox{0.76\linewidth}{!}{
        \begin{tabular}{c|c|c|c|c}
            \toprule
          \multirow{2}{*}{\textrm{\textbf{LongBench}}} &  \(\mathbf{(K16,V16)}\) & \(\mathbf{(K16,V4)}\) & \(\mathbf{(K16,V2)}\) & \(\mathbf{(K2,V16)}\)  \\
          \cmidrule{2-5}
          & 49.26 & 49.54 & 49.30 & 47.73 \\
            \bottomrule
        \end{tabular}
    }
\end{table*}

\section{Limitation}
\label{sec:limitation}
Although \name\ achieves promising results in reducing storage and
computational resources, we also discuss the limitations of our current work.
This work focuses exclusively on decoder-only Transformer-based large language models (LLMs) that utilize rotary position embedding (RoPE) as the underlying position encoding mechanism. 
RoPE has become a prevalent choice in many state-of-the-art open-source LLMs. 
However, the effectiveness of \name\ when applied to models with alternative position encoding methods or attention mechanisms remains an open question and warrants further investigation.
Furthermore, exploring more recent LLM backbones~\citep{qwen3} is necessary, but due to time and computational limitations, we leave this part of the work for future research.

\end{document}